\documentclass[journal]{IEEEtran}
\ifCLASSINFOpdf
\else
\fi
\usepackage{amsmath,graphicx,url,algorithm,algorithm,amsfonts,epstopdf,flushend}
\usepackage[noend]{algpseudocode}

\begin{document}
\title{Partially Occluded Leaf Recognition via Subgraph Matching and Energy Optimization}

\author{Ayan~Chaudhury
        and~John~L.~Barron
\thanks{The Authors are with the Department of Computer Science, University of
Western Ontario, London, Ontario, Canada. Email: \em \{achaud29,barron\}@csd.uwo.ca}}

\markboth{}%
{Shell \MakeLowercase{\textit{et al.}}: Bare Demo of IEEEtran.cls for Journals}

\maketitle

\begin{abstract}
We present an approach to match partially occluded plant leaves with databases of
full plant leaves. Although contour based 2D shape matching has been studied extensively 
in the last couple of decades, matching occluded leaves with full leaf databases is an 
open and little worked on problem. 
Classifying occluded plant leaves is even more challenging than full leaf matching
because of large variations and complexity of leaf structures.
Matching an occluded contour with all the full contours
in a database is an NP-hard problem [Su et al. ICCV2015], so our algorithm
is necessarily suboptimal.

First, we represent the 2D contour points as a $\beta$-Spline curve. 
We extract interest points on these curves via the
Discrete Contour Evolution (DCE) algorithm. To find the best match of an occluded curve
with segments of the full leaf curves in the database, we formulate our solution 
as a subgraph matching
algorithm, using the feature points as graph nodes. This algorithm produces one or more
open curves for each closed leaf contour considered. These open curves
are matchable, to some degree, with the occluded curve. 
We then compute the affine parameters for each open curve
and the occluded curve. After performing the inverse affine
transform on the occluded curve, which allows the occluded curve and any subgraph curve 
to be ``overlaid'', we then compare the shapes using the Fr\'echet distance metric.
We keep the best $\eta$ matched curves.
Since the Fr\'echet distance metric is cheap to compute but not perfectly correlated with
the quality of the match, we formulate an energy functional that is well correlated
with the quality of the match, but is considerably more expensive to compute.
The functional uses local and global curvature, angular information and local geometric features. 
We minimize this energy functional
using the well known convex-concave relaxation technique. The curve
among the best $\eta$ curves retained.
that has the minimum energy, is considered to be the best overall match with the occluded
leaf. Experiments on three publicly available leaf image database shows that our
method is both effective and efficient, outperforming other current state-of-the-art methods.
Occlusion is measured as a percentage of the overall contour (and not leaf area) that is missing.
We show our algorithm can, for leaves that are up to 50\% occluded,
still identify the best full leaf match from the databases.
\end{abstract}

\begin{IEEEkeywords}
Leaf Classification, Beta Splines, Affine transformations, Curvature, Shape Context,
Subgraph Matching, Fr\'echet Distance, Energy Optimization.
\end{IEEEkeywords}

\IEEEpeerreviewmaketitle

\section{Introduction}
\IEEEPARstart{S}{hape} matching is a long standing problem in Computer Vision and
Pattern Recognition. Shapes can appear in different ways, such as discrete point 
clouds, triangular surfaces and silhouettes. 
Recognition of an object by it's contour has been shown to be feasible
for a large variety of shapes \cite{closedContour-CVPR-2000}.
Since the contour of an object is not
affected by colour, illumination or other textural properties, a contour
based representation can be very effective as a shape representation scheme.
The recognition problem can be solved by either correspondence matching between
two point sets, or as a part matching problem by ``learning" object parts from a database
and then recognizing an object from the learned categories.
In the former case, a typical approach is to find good feature points
(or descriptors encoding the local and global geometrical structures of the shape)
on the contour and then to match the feature points between two images to find
the correspondence (\cite{partialshape-MCMC-CVPR-2011}, \cite{shapeLandmark-CVPR-2004}).
The quality of a match is typically determined by a ``score" function on the
matching. Another
common approach to find the point correspondence is by formulating
the point matching as an assignment problem and optimizing it.
Shape Context (SC) \cite{SC-PAMI-2002} is a well known example of such a technique, which builds a 
log polar histogram for each contour point and finds the best match by formulating
the matching as
a bipartite graph matching problem that minimizes a cost function. The method is 
state-of-the-art and works well on large varieties of datasets.  
Scott \emph{et al.} \cite{contourAssgnmnt-TIP-2006} presented a contour matching technique
that improved on the typical assignment methodology by considering the ordering of the discrete
contour points. Ling \emph{et al.} \cite{IDSC-PAMI-2007} extended the idea of shape
context by replacing the
Euclidean distance by a inner distance metric, called the Inner
Distance Shape Context (IDSC). They reported much higher 
recognition rates than SC by using a Dynamic Programming approach.
Later the idea of shape context was used in various applications. 
Ma \emph{et al.} \cite{shapesimilarity-CVPR-2011}
proposed a technique to match partially occluded object contours.
They used a shape context based descriptor and formulated partial graph matching as
a weighted graph matching problem. Also, there has been attempts to handle local deformation
in the shape contour. Xu \emph{et al.} \cite{contourFlexibility-PAMI-2009}
defined a deformable potential at each point on the contour, which can
handle deformations. 
Kontschieder \emph{et al.} \cite{BeyondPairwise-ACCV-2009}
formulated a \emph{k}-nearest neighbour graph for shape representation and
performed an unsupervised clustering method to retrieve the closest match of a
query shape. Hu \emph{et al.} \cite{angularShape-TIP-2014}
proposed a shape descriptor based on angular pattern of contour points.
The descriptor is invariant to scale and rotation due to the relative orientation
of the contour points.

Another approach for shape recognition involves part learning. The idea is
to build a ``dictionary" of object parts and then find the best match
from the learned object parts. Opelt \emph{et al.}
\cite{boundaryFragment-ECCV-2006} encoded contour 
fragments into a weak classifier and formulated a ``boundary fragment model"
to classify an object. They perform boosting, which learns from
a number of weak classifiers and ``boosts" these classifiers to form a strong classifier. 
One advantage of the algorithm is that, it needs a limited number
of training samples. A similar approach was used by Shotton \emph{et al.}
\cite{contourFragment-PAMI-2008}. They built a ``codebook" of local
fragments of contours and learned via a boosting algorithm. 
Ferrari \emph{et al.} \cite{segmentNetwork-ECCV-2006} demonstrated a rule based
approach to recognize objects by their parts. Different parts of an object contour are 
connected to form a contour segment network and matching is performed by
finding efficient paths through the network.
Bai \emph{et al.} \cite{vocabularyShape-TIP-2014}
presented a descriptor learning approach by exploiting the 
``bag-of-words" idea.
A spatial pyramid based approach is adopted to learn local and global 
features at different levels.

Apart from matching contours, either via pairwise point matching or part
learning, other approaches have been shown to be successful to some extent.
In many cases, an object can be efficiently represented by it's meaningful
parts. Although it's hard to define ``meaningful" quantitatively, 
convexity and curvature information can be useful since these
are strong cues for human vision. Discrete Contour Evolution (DCE) \cite{DCE-CVIU-1999}
is a popular technique to decompose an object into meaningful parts based
on the convexity of the shape contour.
Another way to recognize meaningful features is to find the points
where the curvature undergoes change in sign.
Mokhtarian \emph{et al.}'s \cite{CSS-BMVC-1996}
Curvature Scale Space (CSS) feature matching technique has been quite successful
for closed contour matching \cite{closedContour-TCSVT-2004}.
They detect feature points at different scales based on curvature sign change
and match those features to find the shape similarity. Wang \emph{et al.}
\cite{splineSmooth-PAMI-2007} used this idea by representing the shape contour
as B-spline curve. The idea of CSS was later extended for open curves
\cite{CSS-curve-ICIP-2013}. Topological skeleton (medial axis) based approaches (\emph{shock graph}
\cite{shockGraph-PAMI-2004}) are dependent on the skeleton of a curve being well represented
by its skeleton. This works well for shapes with large variations. However, sometimes
small local changes in the shape cause
large change in the skeleton. For leaf shape recognition, with large local shape variations,
this type of idea would not work.

Curve moments are shown to be effective for many cases of contour matching 
\cite{affineMoment-PR-1997}. Representing the contour by a B-spline and  matching 
the spline curves by curve moments is reported to work well for occlusion
handling and missing data (\cite{affine-TIP-1995}, \cite{affine-TIP-1996}). 
However, the method can't be applied to real time applications because the 
matching needs to be performed for all possible combinations of discrete curve
sections in the database. Also, curve moments are very unstable, as
small changes in shape can cause significant changes in the curve moment.

This paper considers the matching of full plant leaves with partially occluded leaves.
We assume that the contour of the occluded leaves is known.\footnote{Full leaf contours are closed boundaries while occluded leavers are open boundaries.}
Matching plant leaves is extremely challenging
due to large intra-class and minor inter-class variations. 
Although the above mentioned methods work well on standard shape databases, 
they have limited success in handling occlusions.
When the occlusion level
is high, most of the methods fail to produce good recognition rates.
Partial shape matching has been studied explicitly in the literature
(\cite{partialWaterman-CVPRW-2008}, \cite{partial-ACCV-2009}),
but extensive experiments on occluded shapes have not been reported. 
To the best of our knowledge, modelling occlusions, especially for leaf images has not
been studied before. We believe that
we are the first to present a method to recognize plant leaves
when the tested occlusion level is as high as $50\%$. Note that it is not the 
percentage of the occluded leaf that is available for matching
that matters but the amount of ``structure'' the occluded leaf has for matching purposes.
Therefore, other leaf species could support more or less occlusion matching than we achieved 
for our leaf datasets.

In next section we discuss related work on leaf matching.
Then we state our main contribution and give an overview of our algorithm.
Then we present our approach, followed by experimental
results and conclusions.

\section{Related Work}

Automatic identification of plant species is an active area of research these
days. There are many ways to identify a plant, for example, by the flowers, fruit,
leaves, or other organs of the plant. Although there has been some work
on flower classification \cite{flowerClassification-CVPR-2006} and
large scale plant identification from learned categories \cite{treeClassifier-TIP-2015},
classifying leaves in order to determine the species is the most common approach.
There is a large body of literature (\cite{leaf-intersection-TIP-2004,leafVenation-CVIU-2008,
MDM-TIP-2012,leafsnap-ECCV-2012,leafMargin-VISAPP-2013,leafMobile-ICIP-2013,
treeIdentification-CVIU-2013,compoundLeaf-ICIP-2013,leafSalient-ICIP-2013,
leafMargin-PRL-2014,leafSpecies-ICPR-2014,stringCut-TIP-2014,
leafCounting-PR-2015,leafMobile-InfoSc-2015,
leafTooth-Plos-2015,leafSalient-ICIP-2015}) over the
last two decades on leaf recognition for plant species identification.
Leaf classification was first reported by Mokhtarian \emph{et al.}
\cite{leaf-intersection-TIP-2004}. They studied recognition of leaf species having
self intersections by Curvature Scale Space (CSS) matching. However, they focused on
the particular case of self intersecting leaf images and did not report 
recognition for normal leaves. Nam \emph{et al.} \cite{leafVenation-CVIU-2008}
performed a nearest neighbour based feature point matching technique to
recognize leaves. Recently, Multiscale Distance Matrix (MDM) \cite{MDM-TIP-2012}
has been a popular leaf recognition technique. The method
is simple and fast. Basically, pairwise distances among all discrete contour
points are represented in matrix form and the course to fine level details
of the shape is encoded by a compact representation of the distance matrix.
This matrix is used to perform recognition by reducing the dimensionality,
similar to a principal component analysis. The method achieves good performance on 
two leaf databases. However, they did not study occlusion. 

In recent years, one of the most successful and practical leaf recognition systems
is Leafsnap \cite{leafsnap-ECCV-2012}. This mobile app system scans a leaf
image (which needs to be on a  uniform background), segments the image, extracts its
contour and matches the contour with thousands of leaves in the database. The underlying 
method computes the Histogram of Curvature over Scale (HoCS) and performs matching via
a nearest neighbour technique. They did not consider occlusion. 
Curvature Scale Space \cite{leafMargin-VISAPP-2013} and  convexity based methods
\cite{leafMobile-ICIP-2013} are also known to
have some success in leaf recognition.
Cerutti \emph{et al.}'s \cite{treeIdentification-CVIU-2013} method focuses 
on the segmentation of a leaf from it's complex background and ultimately performs 
matching based on a CSS formulation. However, the method is dependent on
sharp features (like leaf teeth and tip) on the leaf contour. 
Identification of compound leaves\footnote{A plant leaf consisting of a number of distinct
parts (leaflets) joined to a single stem} is more challenging than classifying
individual normal leaves \cite{compoundLeaf-ICIP-2013} and is not considered here.

Some techniques exploit salient points on the leaf contour
(\cite{leafSalient-ICIP-2013}, \cite{leafMargin-PRL-2014},
\cite{leafSalient-ICIP-2015},  \cite{leafTooth-Plos-2015}), and matching is
performed based on those features. Nevertheless, this idea fails when the
leaf does not have enough distinguishable features. 
Hierarchical representation of the shape contour is an efficient way to capture
the geometry at different scales, which can be exploited using a Shape Tree (ST)
\cite{shapeTree-CVPR-2007}. Another hierarchical approach for partitioning the contour 
into different lengths was presented by Wang \emph{et al.} \cite{stringCut-TIP-2014}.
They captured the geometrical structure via the
distribution of points around a straight line and then built a shape signature
based on Fourier transform coefficients. The method achieves good performance 
on a public database of Swedish leaves, including leaves that are collected
from $100$ tree species. 
Some recent work (\cite{leafSpecies-ICPR-2014}, \cite{leafMobile-InfoSc-2015})
developed leaf recognition systems for mobile applications. 
Recently Zhao \emph{et al.} \cite{leafCounting-PR-2015} demonstrated a 
computationally fast technique for leaf recognition. However, the algorithm requires prior
training for the classifier. Furthermore, occlusion is not handled by 
their method. 

\section{Contribution}

This paper explicitly deals with matching occluded partial leafs with 
unoccluded full leaves. The rate of
occlusion is defined as the percentage of the contour that is missing
(say, occluded by other leaves or objects). 
Currently, only one occlusion event per occluded leaf can be handled.
The task is to find partial contours in full leaf contours that match the
occluded contour. Other papers sometimes say they are robust to occlusion 
but offer no evidence 
(\cite{partial-ACCV-2009,shape-ACCV-2014,hmm-TIP-2007,partialWaterman-CVPRW-2008})
or offer evidence for small occlusions only (about $10\%$)
(\cite{shapeWoCorr-PAMI-2012,shapeSoftClustering-PAMI-2012}). 
These later papers do not explicitly model occlusion but treat occlusion as measurement error.

Our algorithm offers a sub-optimal solution to the general partial contour 
matching problem that is known to be NP-Hard \cite{contour-NP-hard-ICCV-2015}.
Our algorithm uses many tools available in the literature, for example,
$\beta$-splines, the Discrete Contour Evolution algorithm, subgraph matching, 
affine transformations, the Fr\'echet distance metric and the
GNCPP convex-concave relaxation optimization method. However, it is
the algorithm using these tools that is novel here.

\section{Algorithm Overview}

Our algorithm represents the contours of leaves
with spline curves (which allows arbitrary curve resolution) and extracts feature points (for 
example, points with high convexity) from the curves. Then, we use subgraph matching 
to match an occluded leaf curve to possible partial curves belonging to full leaf curves. 
This method yields a number of partial curves 
from the full curve. We can compute the affine transformation for the occluded curve
and each partial curve and then
overlay the occluded curve with each partial curve (the affine transformation 
is shape preserving). A small number of the best matches of the overlaid curves 
according to the Fr\'echet distance metric are retained. These matches are ``globally good"
matches in that the overall shape of the matched curves is almost identical.
It is still possible for any two curves to have a low Fr\'echet error and still
not be a good match because of local structure differences.

To handle the finer local structure differences of a small subset of the
curves with the best Fr\'echet scores, we formulate an energy functional based on 
local and global curvature and the angular and geometric features of the curves. 
Optimization of this energy functional gives the best match for the occluded curve
and the best partial curves of full leaves (according to the Fr\'echet scores), 
where the local curve structure is now taken 
into account. Fr\'echet matching is fast while optimization of the energy functional is expensive.
Our algorithm leverages the advantages and disadvantages of these two matching techniques.

\section{Our Approach}

Splines are powerful tools for representing a curve mathematically. Among the different types of
spline curves, B-spline based contour representations has been successfully used for 2D
shape matching (\cite{splineMDL-TIP-2000}, \cite{splineSmooth-PAMI-2007},
\cite{superCurve-TIP-2004}).
Due to its smoothness and continuity properties, B-splines are extremely useful in
approximating the boundary of an object. B-spline curves are piecewise polynomial functions
where local curve approximation is performed using control points. Local control is extremely
useful for modelling a contour to a desired level of detail. Another interesting
property of B-spline is that it is affine invariant. This property is
very useful when matching two curves which
are related to each other by an affine transformation.

In general, a $(p+1)$-th order (i.e. $p$-th degree) B-spline is $C^p$ continuous.
A B-spline curve $P(u)$ defined by $(n+1)$ control points
$P_0, P_1, \cdots, P_n$ is defined as:
\begin{equation}
P(u) = \sum_{i=\lfloor nu-2 \rfloor}^{\lceil nu+2 \rceil} B(nu-i) P_i,
\label{eqn_b_spline}
\end{equation}
\noindent where $B(nu-i)$ is the blending function of the spline
(a bell shaped curve is non-zero when the inequality $-2 < nu-i < 2$ holds) and $u \in [0,1]$.
[Equation (\ref{beta-eqn}) below with $s=1$ and $t=0$ gives this blending function.]
In case of a cubic B-spline, at most four nearest control points are used to compute the
blending function for a spline point (if $u=i/n$ then only 3 control points are needed).

For any set of 2D control points $P_i = (x_i,y_i)$, we can always obtain the
spline point $(x(u), y(u))$ for any $u \in [0,1]$. Thus the number of spline points can be greater than, 
equal to or less than the number of control points.
Any curve thus can be represented as the polyline joining continuous set of spline points.

\subsection{$\beta$-Spline Based Representation}

Due to computational efficiency, cubic B-splines are typically used to model a 
contour \cite{BurgerGillies}.
Cubic B splines are good for approximating the curve whose shape is controlled by
the control points. However, to interpolate the control points, one needs to solve linear
systems of equations to solve for some {\it phantom} control points, such that the 
approximation of these phantom control points interpolates the real control points.
B-spline interpolation is computationally expensive and complex. $\beta$-splines
(\cite{betaspline-1985-Goodman,betaspline-1986-Goodman}) provide an intermediate 
representation between approximation and interpolation by providing a tension parameter, $t$. 
$t=0$ yields a B-spline while $t > 0$ (say, $t \in [3,15]$) provides 
a spline that almost perfectly interpolates the control points. [A second parameter, skew $(s)$,
causes undesirable discontinuities in the spline curve and is usually ignored, 
i.e. we keep $s=1$.]
Like B-splines, $\beta$-splines are $1^{st}$ and $2^{nd}$ order continuous and require at
most 4 control points for the computation of a spline point value $P(u)$, $u \in [0,1]$.

Other approaches to produce smooth contour curves include Thin Plate Splines
(TPS) \cite{TPS-CVIU-2003} or Relevance Vector Machine (RVM) regression
techniques \cite{RVM-2012}. These techniques probably could be used in place
of $\beta$ splines with good results being obtained. We chose $\beta$-splines for simplicity
and efficacy reasons.

Having $(n+1)$ control points, $P_0, P_1, \cdots, P_n$, where the original contour points
are used as the control points, a $\beta$-spline can be computed as:
\begin{equation}
P(u) = \sum_{i=\lfloor nu-2 \rfloor}^{\lceil nu+2 \rceil} \beta (nu-i) P_i,
\label{eqn_beta_spline}
\end{equation}
where $u \in [0,1]$. Note that the  minimum and maximum bounds on $i$ in 
Equation (\ref{eqn_beta_spline}) ensure that
the inequality $-2 < nu-i < 2$ is satisfied and the blending function $\beta$ is always 
non-zero for this range of $nu-i$ values. [All other $i$ values yield $nu-i$ values 
outside this range and $\beta$ is always 0 for these values. 
Note that the loop specified in Equation (\ref{eqn_beta_spline})
could yield a few $\beta$ values being zero because of the 
floor and ceiling functions used in the calculation of $i$ but no non-zero
$\beta$ values are ever missed.]
The blending function $\beta(\tau)$ is
given by (\cite{betaspline-Barsky1983,BurgerGillies}) as:\\
\begin{equation}
\beta(\tau) = 
\begin{cases}
\frac{2}{\delta}(2+\tau)^3, \hspace{38mm} -2 \leq \tau \leq -1 \\
\\
\frac{1}{\delta}((t+4s+4s^2)-6(1-s^2)\tau - 3\tau^2(2+t+2s) \\
-2\tau^3(1+t+s+s^2)), \hspace{22mm} -1 \leq \tau \leq 0 \\
\\
\frac{1}{\delta}((t+4s+4s^2)-6\tau(s-s^3) -3\tau^2(t+2s^2+ \\
2s^3) + 2\tau^3(t+s+s^2+s^3)), \hspace{15mm} 0 \leq \tau \leq 1 \\
\\
\frac{2}{\delta}s^3(2-\tau)^3, \hspace{39mm} 1 \leq \tau \leq 2,
\end{cases}
\label{beta-eqn}
\end{equation}
\noindent where $\tau = nu-i$,
$t$ is the tension parameter and $s$ is the skew parameter.
$\delta$ is given by:
\begin{equation*}
\delta = t+2s^3+4s^2+4s+2.
\end{equation*}

For our purpose, skew is not a useful parameter (it can introduce discontinuities into
the spline curve) and we always use $s=1$.
Like cubic B-splines, the blending function is still symmetrical with respect to $\tau$, 
which implies $\beta(-\tau) = \beta(\tau)$. 
Also, the sum of all the non-zero $\beta$ values is always 1:
\begin{equation}
\beta(\tau-2) + \beta(\tau-1) + \beta(\tau) + \beta(\tau+1) = 1.
\end{equation}

A $\beta$-spline is second order continuous everywhere when skew $s=1$.
For tension $t=0$ and skew $s=1$, the $\beta$-spline reduces to a cubic B-spline.
Positive values of tension $(t>0)$ increase the amplitude of the
two middle segments of the third-order curve with respect to the first and 
last segments, whereas increasing the skew $(s>1)$ increases the amplitude of the
two segments on the right with respect to the two segments to the left side of the curve and 
so may produce discontinuities and other unwanted artifacts.
We used $t=10$ and obtained experimental results that were typically 5\%-10\% better 
than if we used B-splines only ($t=0$).

\section{Approximate Curve Section From Full Leaf}

Once we have the $\beta$-spline representation of the leaf contour, we perform
curve matching as discussed below.
For the occluded test leaf, we need to determine the closest match of an occluded curve
to all curve sections of all the full leaves in the database. 
Matching two curves is a well known problem and has been well studied in Computer Vision.
For matching of curves representing leaf contours,
the problem is even harder than the general case due to several factors. There are many 
intra-class variations for many leaf species. Moreover, the boundary of the 
leaf contour relative to the background or to other leaves
might not be smooth due to segmentation errors. In our case,
the curve matching problem is complicated because we have to determine
which part of the full curve matches the occluded curve. One idea could be to use
a brute force approach to consider every possible combinations of discrete
points in the full curve and find the match with the occluded curve. This
is a NP-hard problem \cite{contour-NP-hard-ICCV-2015} that can't be solved realistically
for any reasonable number of contour points. 
We handle the problem in the following way. 

First, we smooth the contour using the Savitzky-Golay filter. The idea 
is to fit a $2^{nd}$ order polynomial to the data points via least squares.
More specifically, we apply this filter by using the \emph{smooth()} function in
the Matlab curve fitting toolbox. The purpose of smoothing
is to eliminate local noise and segmentation error at the boundary of the leaf.
However, a small neighbourhood should be 
chosen to perform the smoothing operation, otherwise the smoothing may
suppress useful local geometrical structure of the leaf. 
With the smoothed curve sections, we next perform feature detection and then
match the features of the two curves. 

\begin{figure}
\includegraphics[width=3.0in,height=2.0in]{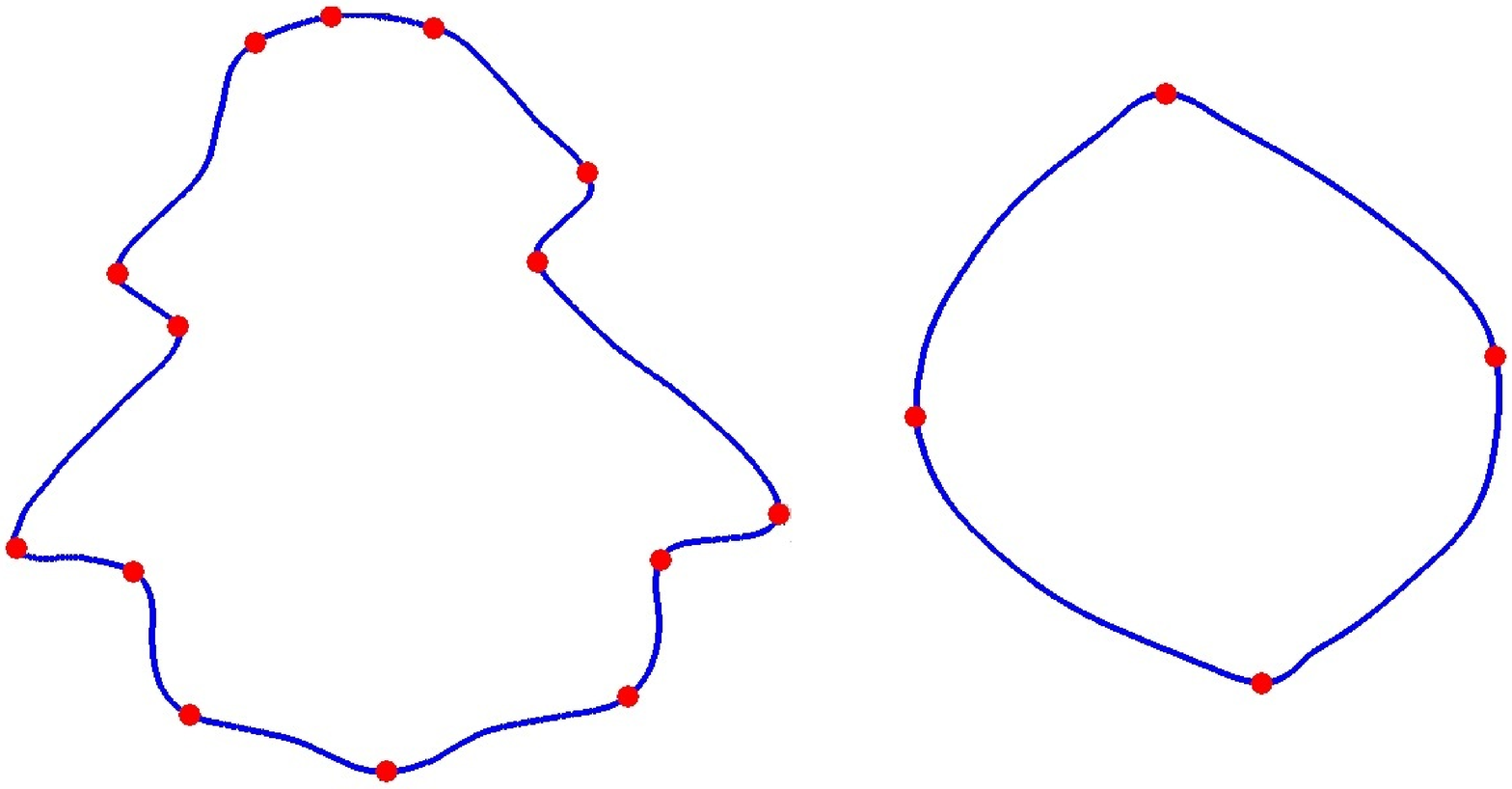}
\caption{Feature points obtained from Discrete Contour Evolution (DCE) method
\cite{DCE-CVIU-1999}. The contour
is decomposed into meaningful parts based on the convexity and the feature points
represent the joint of these parts (or the vertices of the polygon represent the
shape).}
\label{dce_ex}
\end{figure}

Although global shape descriptors have been shown to have some success in shape matching,
the idea mostly works for shapes with large variations. Leaves not having unique
distinguishing properties cannot be matched with traditional shape descriptors.
Convexity is an important cue in human vision, and shapes can be decomposed into
meaningful parts using convexity information. We use the Discrete Contour Evolution (DCE)
\cite{DCE-CVIU-1999} to find interest points on the 2D contour of the leaf. 
The idea of DCE is to simplify the contour by hierarchically decomposing the boundary
by representing the shape with a polygon. The vertices of the polygon 
represent convex parts of the shape. Figure \ref{dce_ex} shows examples of DCE
features. The first leaf contour consists of a lot of variations, whereas the
second leaf contour lacks any meaningful features on the contour. One advantage of using
DCE feature points is that, the number of features is invariant with respect to scale. 
Thus, two similar leaves of different sizes should have similar feature point
pattern in their contours. 

With the DCE feature points in the occluded and full contour, we next need to
determine what is the best match between two set of feature points. We
formulate this problem as a subgraph matching optimization problem.

\subsection{Subgraph Matching}

Treating the feature points
as graph nodes, the problem can be formulated as a subgraph isomorphism problem. 
Two graphs $G_1=(E_1,V_1)$ and $G_2=(E_2,V_2)$ are isomorphic, denoted by 
$G_1 \cong G_2$, if there is a bijection $\varphi: V_1 \rightarrow V_2$ such that
for every pair of vertices $v_i,v_j \in V_1$, edge $(v_i,v_j) \in E_1$ if and only if
$(\varphi(v_i),\varphi(v_j)) \in E_2$.
Although there are many exact isometric matching
algorithms, our case is more complicated for several reasons. First,
we need to deal with missing nodes. Graph nodes can be missing due to 
inconsistencies between the detected feature points for two similar leaves, 
which may be slightly different in local geometry.
Second, we have to consider the cases where $G_1 \cong G_2$
or $G_2 \cong G_1$ because the number of feature points in the occluded curve
can be less than the number of feature points of the full curve, and vice versa.


\begin{figure}[ht!]
\includegraphics[width=3.5in,height=2.0in]{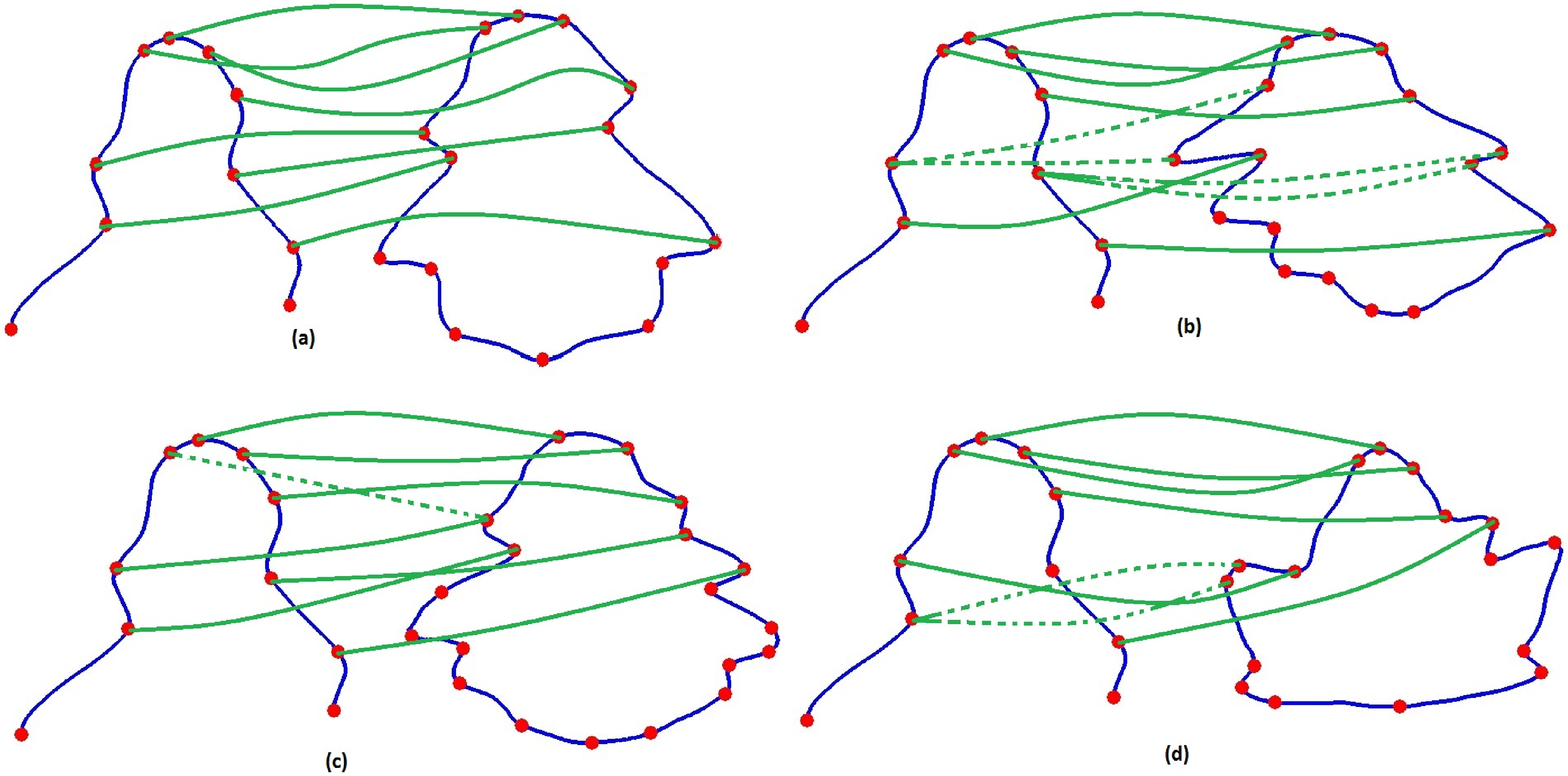}
\caption{Four examples of graph matching. An occluded curve section is matched
with $4$ full curves in the database (the left curve is the occluded curve and the
right curve is the full curve).
In (a), (b) and (c), an occluded curve section is matched with
full curves from the same species. In (d), the same occluded curve
is matched with a different species.
Good matches are shown as solid green lines while ambiguous matches are shown as
dotted green lines.}
\label{graphmatch}
\end{figure}

Figure \ref{graphmatch} shows a few examples for
matching an occluded curve section with some full curves in the database.
In each of the four examples, the left curves are occluded curves
and the right curves are the full curves.
First, we extract the feature points on the contour via DCE as discussed previously. 
Then we build a connected graph with all the feature points as nodes and
the geodesic distances as edges. 
In Figures \ref{graphmatch}a, \ref{graphmatch}b and \ref{graphmatch}c,
the full curves are from the same
species, but in Figure \ref{graphmatch}d the full curve is from a different
species. In each case,
we want to find a match with an open curve section (or an occluded leaf).
In the first three cases we have shown how the ambiguities in graph match can
occur within a single species. Figure \ref{graphmatch}a shows a
good match. Although
it's not ``exact", the overall topologies are similar in the occluded curve
and the section of the full curve under consideration. Matches are shown
in green lines. Now consider Figures \ref{graphmatch}b and \ref{graphmatch}c.
Although these curves are also
from the same species as in Figure \ref{graphmatch}a, there are matching ambiguities
in the graph nodes. Matches with no ambiguities are shown as solid green lines while
ambiguous matches are shown as dotted green lines. This is because one node
in the occluded curve can match multiple nodes in the full curve and
vice versa. Finally in Figure \ref{graphmatch}d,
the occluded curve section is matched with a full curve from a different
species. The vertices in the occluded curve still find matches with the
vertices in the full curve, and there are some ambiguities in the matches.
From the examples in Figure \ref{graphmatch}, we can infer that there is a
need to quantify the quality of match by a ``score''. Also, we need to handle
the cases of ambiguities of unmatched vertices.

We base our ``scoring'' function on the ideas adopted from 
Mcauley \emph{et al.} \cite{occludedGraphMatch-PR-2012} graph matching.
For two graphs $G_1$ and $G_2$, we 
assume that outliers can be present in either of the graphs. We wish to find
a function $\hat{f}: G_1 \rightarrow G_2$ such that the distances
between the points in $G_1$ and $G_2$ are minimized. Along with a
penalty term for unmapped points between the two graphs due to occlusion
and missing data, the graph matching problem can be formulated as the
following energy function:

\begin{equation}
\begin{split}
\hat{f} = \arg\min_{f: G_1 \rightarrow G_2} \sum_{(i,j) \in G_1}
|d(g_i,g_j) - d(f(g_i),f(g_j))| + \\
\lambda \underbrace{(|G_1| - |f(G_1)|)}_{\text{number of unmapped points}},
\end{split}
\end{equation}

\noindent where $d$ is the geodesic distance between the nodes and 
$\lambda$ is the maximum
number of outliers that are likely to be present. We use the graph topology 
as proposed by Mcauley \emph{et al.} \cite{occludedGraphMatch-PR-2012}
and use two nodes $g_1$ and $g_2$
as reference nodes. We order the vertices in counter clockwise
direction and choose the first and last nodes as reference nodes. First we find
the optimal mapping of two point sets using Algorithm \ref{subgraph}.
Then we repeat the process by selecting every pair of nodes as reference
nodes and find the optimal cost, which gives the optimal correspondence. 
After finding the best matched feature points between the two graphs, we
retrieve the corresponding curve section from the full curve.

\begin{algorithm}
\caption{Subgraph matching algorithm}\label{subgraph}
\begin{algorithmic}[1]
\State $best \gets \phi$ , $bestcost \gets \infty$ 
\For{each possible mapping $f(g_1),f(g_2) \in G_2 \times G_2$}
\For{$r \in \{1,-1\}$}
\State $cost \gets ||g_1-g_2| - |f(g_1)-f(g_2)||$
\For{for each node $g_i \in G_1 \backslash \{g_1,g_2\}$}
\State Find expected position $p$ of $f(g_i)$
\State Find $p$'s nearest neighbour $n_p$
\State $cost = cost + ||g_1-g_i|-|f(g_1)-n_p|| + ||g_2- 
g_i|-|f(g_2)-n_p||$
\If{$cost > bestcost$}
\State break
\EndIf
\EndFor
\If{$cost < bestcost$}
\State $best \gets f$
\State $bestcost \gets cost$
\EndIf
\EndFor
\EndFor
\State \textbf{return} $best$, $bestcost$
\end{algorithmic}
\end{algorithm}

So far, we have extracted possible curve sections from
all the full curves in the database. Because we perform the matching
for all curves independently, it is unlikely to miss a 
potential match in the process. However, we still need to find
the best match within hundreds or thousands of curves (depending on the size
of the database). The idea is to use a two-stage technique to
reduce the search space. Fist, we want to filter out the 
curves which are too different from each other in terms of global structure.
This filtering is performed by first recovering the global affine transform parameters
between any two curves. Then the
Fr\'echet distance metric is used to measure the closeness between
any two curves and only the best $\eta$ matches are retained.
In a last step, we perform energy optimization to find the overall best answer from
these closely related curve matches (because, infrequently, the Fr\'echet error
can sometimes incorrectly indicate a good match when the global structures are 
similar but the local structures are different).
We discuss these steps in detail in next subsections.

\subsection{Inverse Affine Transform}

\begin{figure}[htb!]
\begin{center}
\begin{tabular}{c c}
\includegraphics[width=1.5in,height=0.75in]{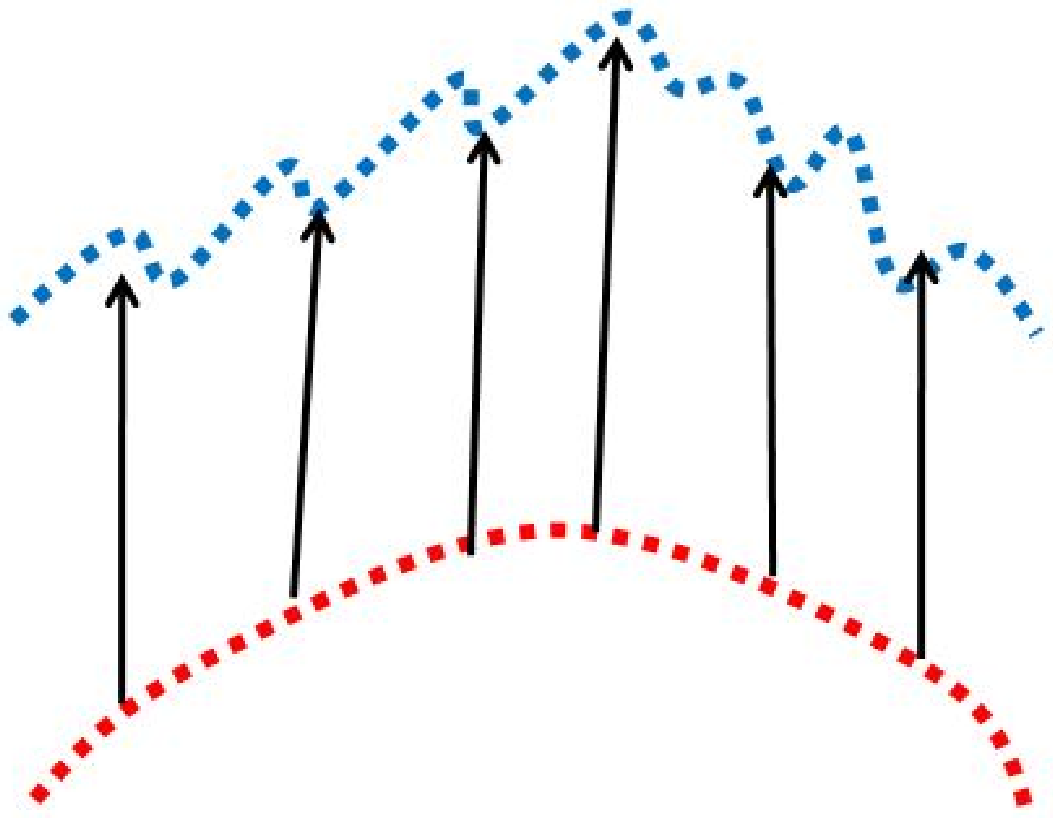} &
\includegraphics[width=1.5in,height=0.75in]{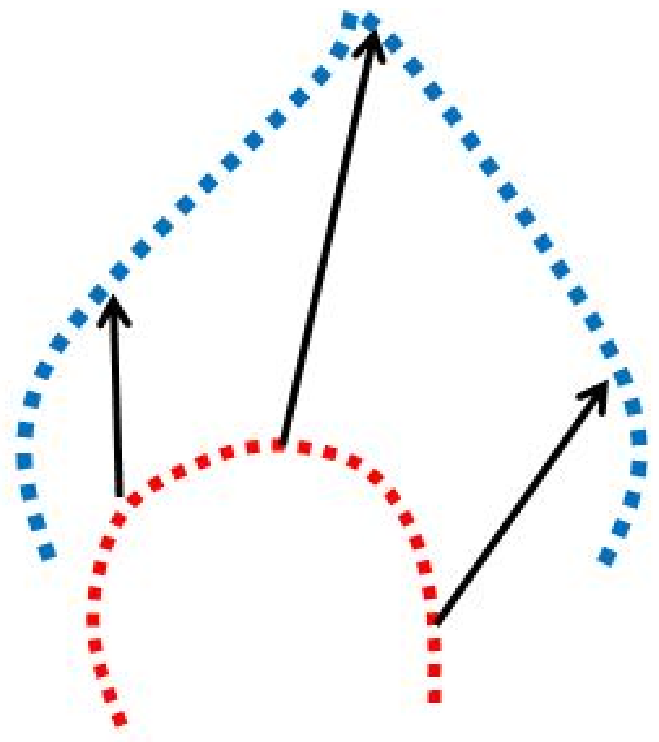} \\
(a) & (b) \\
\includegraphics[width=1.5in,height=0.75in]{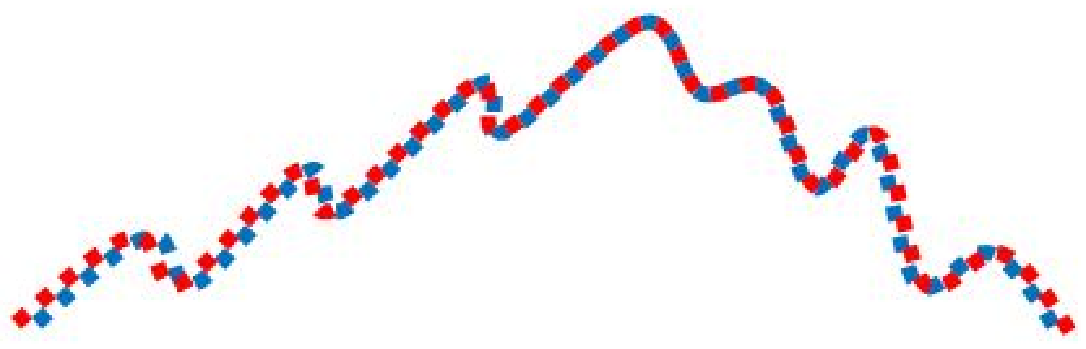} &
\includegraphics[width=1.5in,height=0.75in]{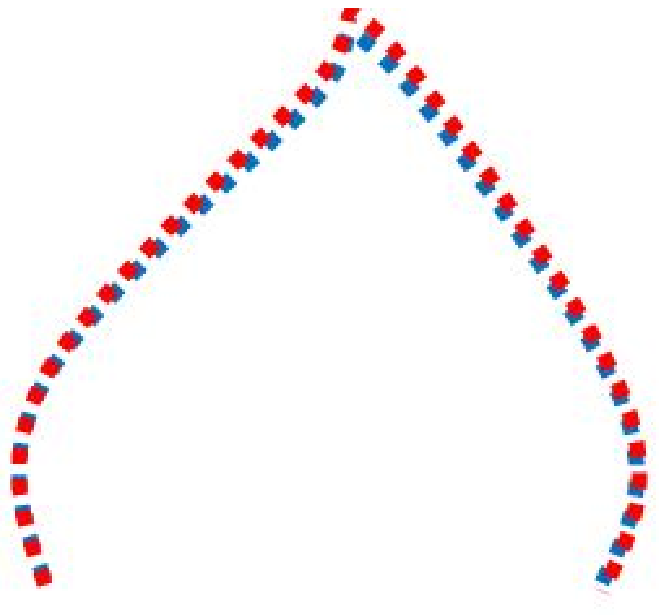} \\
(c) & (d) \\
\includegraphics[width=1.5in,height=0.75in]{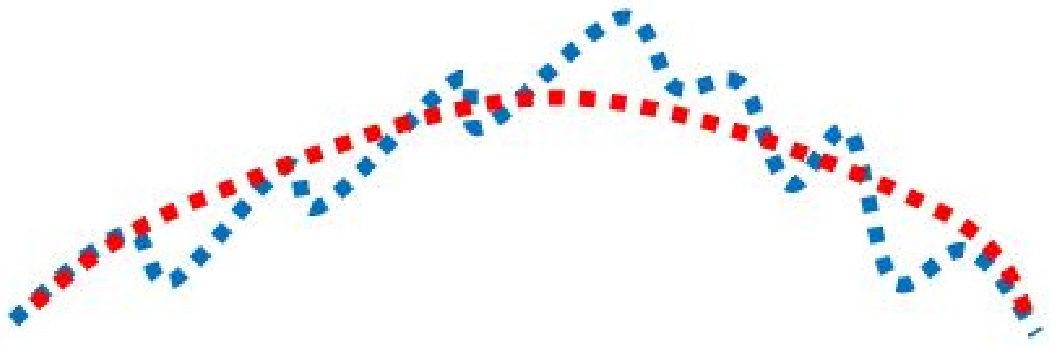} &
\includegraphics[width=1.5in,height=0.75in]{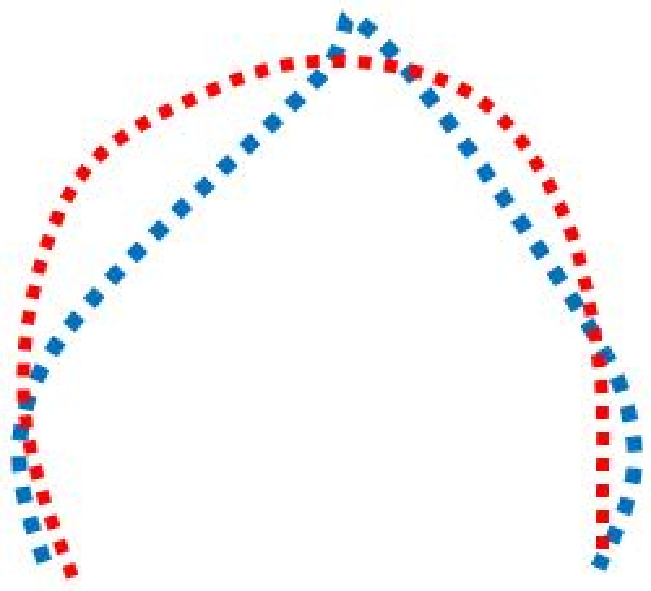} \\
(e) & (f)
\end{tabular}
\end{center}
\vspace*{-0.1in}
\caption{(a) and (b) the correspondence of some points on 2 curves,
(c) and (d) the registration (warping) of the 2 curves and (e) and (f)
the 2 curves overlaid via the computed affine parameters (rotation, translation and scale).}
\label{curve_compare}
\end{figure}

For the same leaf type, the occluded curve and the extracted
curve sections from the full curves are 
assumed to be related by some global affine transformation.
This allows us to recover the global
translation and scale parameters (the rotation is already given
by the graph matching calculation and we haven't encountered shear in
any experimentation on our datasets). Usually, registration papers
present point set registration as warping, where the deformation
parameters needed to warp one set of points into another, are computed.
However, we do not do this, but rather use the affine parameters to
``overlay''  one curve on top of another and then perform a shape similarity
calculation. Figure \ref{curve_compare} illustrates this point.

Figure \ref{curve_compare} show two examples curve matches to emphasize
that we do not to do pairwise registration. The first row of Figure \ref{curve_compare}
shows two point sets from two different curves. The model set is represented in
blue and the data set is represented in red. Some corresponding points are
shown via black arrows in Figures \ref{curve_compare}a and \ref{curve_compare}b.
Note that, since we have modelled the
contour with a $\beta$-spline, we can re-generate exactly same number of
equally spaced points for both spline curves.

If we perform pairwise registration (warping), the algorithm will 
find corresponding points between two curves, and then the warped data set will ``approach" the
model set shape. This transformation is specified using
local and global parameters.
An efficient pairwise registration algorithm will
consider local deformations and will transform each point in the data
set to the nearest point in the model set. This results in something like that
shown in Figures \ref{curve_compare}c and \ref{curve_compare}d. The local structures are treated
as deformations and finally the registration aligns or warps the two point sets to coincide.

We do not model these deformations in our solution because the goal is
to find the  difference between two curves, and not to find the correspondences
between two point sets. Undoing the affine transform between the two curves
does not affect the local structure, it brings the two curves into an
``overlaid" state. Note that, as we have matched feature points via
graph matching and then extracted the curve section from that calculation,
this is automatically rotation invariant. Undoing the affine transform
is a simple but effective way to handle translation and scaling as well
in the global sense.  We compute the affine transform for two curves using the 
MatLab function \emph{fitgeotrans()}.
We show examples of undoing an affine transformation in the third row of the figure, i.e.
in Figures \ref{curve_compare}e and \ref{curve_compare}f.
In these figures, the two curves still have very different structures.

For points $\mathbf{p}_i = (p_{i1},\dots, p_{in})^T$ and their corresponding
affine transformed points $\mathbf{q}_i = (q_{i1},\dots, q_{in})^T$
($i=1,...,m$), the goal is to find the
matrix $\mathbf{A}$ and the translation vector $\mathbf{t}$ such that
\begin{equation}
\mathbf{p}_i \approx \mathbf{A} \mathbf{q}_i + \mathbf{t}.
\end{equation}

\noindent $\mathbf{A}$ incorporates rotation and scaling.
To find $\mathbf{A}$ and $\mathbf{t}$, the following objective function
must be minimized: 

\begin{equation}
S(\mathbf{A},\mathbf{t}) = \sum_{i=1}^m  \left|\left|\mathbf{p}_i - \mathbf{A}\mathbf{q}_i - \mathbf{t} \right|\right|_2^2.
\end{equation}

\subsection{Curve matching by \emph{Fr\'echet Distance}}

After undoing the affine transform, we measure the similarity between the two curves. 
A naive way to do this is simply to sum up the squared difference of all the points from start to 
the end of the curves. However, that does not give a good result.
Typically, the \emph{Hausdorff distance} is used to find similarity between two 
point sets.
By definition, two sets are close in the \emph{Hausdorff} sense if every point of
either set is close to some point of the other set. However, this idea is not effective
for finding the similarity between two curves because it considers only the set of 
points, not their order on the curve. We use the \emph{Fr\'echet distance} 
\cite{Frechet-1995} for measuring the similarity between
two curves. Informally, it can be defined as the following
example\footnote{https://en.wikipedia.org/wiki/Fr\'echet\_distance}:
\emph{Let us suppose that a man
is walking a dog. Assume that the man is walking along one curve and the dog
is along the other curve. Both of them are allowed to control their speed, but
can't go backwards. The Fr\'echet distance between the two curves is the minimum
length of the necessary leash to connect the man and the dog from the beginning till
the end of the two curves}. 

Mathematically, the continuous \emph{Fr\'echet distance}
between two curves, $P$ and $Q$, is defined as:
\begin{equation*}
F(P,Q) = \inf_{\alpha, \beta} \max_{t \in [0,1]}|| P(\alpha(t)) - Q(\beta(t)) ||,
\end{equation*}
where $P,Q: [0,1] \rightarrow \mathbb{R}^2$ are parameterizations of the two curves and
$\alpha,\beta: [0,1] \rightarrow [0,1]$.
In the discrete case, the algorithm can be implemented using dynamic programming
\cite{recurrence-Frechet-2005}. If the curves, $P$ and $Q$, have sizes $m$ and $n$
respectively, then the discrete version of $F(P,Q)$ can be computed by following 
dynamic programming recurrences:
\begin{equation*}
d_{0,0} := ||p_0 - q_0||_2,
\end{equation*}
\begin{equation*}
d_{0,j} := \max \{d_{0,j-1}, ||p_0 - q_j||  \}, ~ \mbox{for j=1:n},
\end{equation*}
\begin{equation*}
d_{i,0} := \max \{d_{i-1,0}, ||p_i - q_0||  \}, ~ \mbox{for i=1:m}
\end{equation*}
and
\begin{equation*}
d_{i,j} := \max \min \{ d_{i,j-1}, d_{i-1,j}, d_{i-1,j-1} \}, ~ \mbox{for i=1:m, j=1:n}.
\end{equation*}

Using the \emph{Fr\'echet distance} metric, we see that similar curves tend to have
smaller distance values. This approach helps to drastically reduce the search
space. However, the approach 
focuses mostly on the global shape of the leaf and can't handle the
cases where the leaves have similar global shape, but have different
local structures. We retain the top $\eta$ matches
from the curve matching
algorithm as discussed above and process these further
as discussed in the next section.

\subsection{Energy Functional}

Now the recognition problem of searching the full database of leaves
reduces to finding the best match
of an occluded curve with $\eta$ curves (which are parts of full curves)
in the database. However, these $\eta$ curves are very close to the test
curve, especially in terms of global structure. To find the best
match among these curves, we need to incorporate both local and
global information to uniquely identify a curve. 
The tunable parameter $\eta$ has value 5 in our work.
This is based on empirical observation
and is used to reduce the search space because energy optimization is 
computationally expensive.
Because the problem of partial contour matching is NP-hard, 
performing energy optimization on the best $\eta$ Fr\'echet matches is
a reasonable suboptimal solution. Otherwise,
a trivial solution of the partial shape matching problem would be
to consider all possible combinations of contour points and find
the best match, i.e. a brute force approach, which has exponential complexity.
However, one can keep more 
than $\eta=5$ matches in the energy optimization phase. This was not needed
for our experimentation with the 3 leaf datasets. For other
applications, $\eta$  might need to be changed.

Local curvature can be useful to encode the local geometric structure of
a leaf. If the leaf does not have much variation (can be thought of
as a ``smooth boundary"), local curvature will have almost the same value at
all points on the contour, except the tip. Leaves having many variations on their boundaries
will have different curvature values at different points. To encode
the global structure of the leaf, global curvature can be an useful
characteristic. However, even if two leaves have similar local and global
structures, one idea is to investigate how they differ in terms of how 
the boundary points are distributed relative to each other.
Another idea for encoding local and global geometrical structure of the
contour is to consider the orientation of other contour points with respect
to a particular point. Relative angles of other points with a particular
point can be used as this metric. 

We formulate an energy
function that involves several geometric features, such as curvature, local and
global geometrical structures, etc. and minimize it to find the best match between two
curves. We formulate the energy functional as:

\begin{eqnarray}
E_{total} & = & \lambda_1 E_{localCurvature} + \lambda_2 E_{globalCurvature} \nonumber  \\
          & + & \lambda_3 E_{angularDistribution} + \lambda_4 E_{stringCut},
\label{energy_eq}
\end{eqnarray}

\noindent where $\lambda_1$, $\lambda_2$, $\lambda_3$ and $\lambda_4$ 
are weight factors for each term. Currently, we set all the weights to $0.25$. 
Basically, we compute four adjacency matrices for the four
terms in the function. For every point on the contour, we compute these factors
for every other point on the contour and encode these in an adjacency matrix.
We explain each term below:

\begin{enumerate}
\item {\bf Local Curvature:} Curvature is an important property for matching shapes. 
Geometrically,  if $\varphi$ is the
angle between the tangent line and the $x$-axis, then the curvature
of a curve $y=f(x)$ is defined as:
\begin{equation}
\kappa = \left| \frac{d\varphi}{ds} \right| = \frac{y''}{[1+(y')^2]^{3/2}}.
\end{equation}
We choose a small number of points ($20$ points in our case) to define the local neighbourhood
around a point of interest and perform the same for all points on the contour
to get local curvature at those points. This information helps
to capture the local geometric structure of the leaf contour. 

\item {\bf Global Curvature:}
This is similar to local curvature, but a bigger neighbourhood is used to capture
the global shape. Also, a bigger neighbourhood is used to smooth the curve while
computing the global curvature, which helps to capture the global geometrical 
structure of the curve. We use $1/3$ of the contour length
to compute global curvature. This captures the overall leaf shape in a global
sense. Note that because we are dealing with open curves 
we use reflection at the end points.

\item {\bf Angular Features:}
As discrete points are uniformly distributed along the contour (because spline points
are uniformly distributed in the interval $u \in [0,1]$), their relative
orientations differ when the shape changes. We considered incorporating relative
angular orientation of the points in our energy function. However, instead of
simply computing the angles, we use Shape Context descriptors
\cite{SC-PAMI-2002}. The idea is to consider a set of vectors originating from
 a point to all other points on the boundary and then compute the distribution 
of all the points over relative positions. For every point $p_i$, a histogram $h_i$
is created in log-polar space, which uses the relative coordinates of all
the points with respect to that point. Mathematically, this is defined in
 \cite{SC-PAMI-2002} as:

\begin{equation*}
h_i(k) = \#\{q \neq p_i : (q-p_i) \in bin(k)\},
\end{equation*}

\noindent where $k$ is the index of the angle-distance bins.
The advantage of using log-polar space is that the descriptor is more sensitive to
the nearby points than to points which are further away. This helps to exploit the
local geometric structure. 

If $p_i$ and $q_j$ are two points on two curves that are to be matched, 
then the cost of matching the curves, computed with the $\chi^2$ statistic, 
can be found as:
\begin{equation*}
C_{ij} \equiv C(p_i,q_j) = \frac{1}{2} \sum_{k=1}^K \frac{[h_i(k) - h_j(k)]^2}{h_i(k) + h_j(k)}.
\end{equation*}

\item {\bf StringCut Features:}
In addition to computing the angular distribution of points in the contour, 
another way to compute local geometric structure is to find the distribution 
of points in a small neighbourhood about a straight line. Inspired by the
work of Wang \emph{et al.} \cite{stringCut-TIP-2014}, we introduce ``StringCut"
features, which contributes to the fourth term of our energy function. 
Figure \ref{stringcut} shows a few examples of local neighbourhoods 
about a point. By drawing a straight line (or ``string") through
the end points (which ``cuts" the set of points), there can be three possible
set of points: the points
on the two sides of the straight line (can be on the above and below sides or, equivalently, on
the left and right sides) and the points lying on the line. 
This idea allows one to extract the local geometry of the neighbourhood.

\begin{figure}
\includegraphics[width=3.5in,height=1.5in]{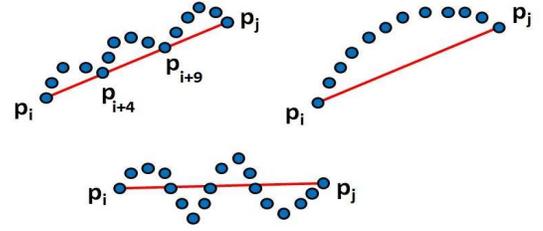}
\caption{The idea of ``StringCut". Some examples of neighbourhoods about 
a point are shown. A straight line is drawn between the first and the last
points of a neighbourhood and the distribution of the points about
the straight line are used as StringCut features.}
\label{stringcut}
\end{figure}

Let a neighbourhood be defined by the points $\{p_i,p_{i+1},\cdots,p_j\}$
which starts from $p_i$ and ends at $p_j$ and the straight line which passes
through the points is denoted by $\xi_{ij}$. The point sets above, below and
on the line are denoted as $S_a$, $S_b$ and $S_o$. For example, in the first 
example of Figure \ref{stringcut}:
\begin{equation*}
\begin{split}
S_a = \{p_{i+1},p_{i+2},p_{i+3},p_{i+5}, \\
p_{i+6},p_{i+7},p_{i+8},p_{i+10},p_{i+11},p_{i+12},p_{i+13}\},
\end{split}
\end{equation*}
\begin{equation*}
S_b = \{\phi\}
\end{equation*}
and
\begin{equation*}
S_o = \{p_i, p_{i+4}, p_{i+9}, p_j\}.
\end{equation*}

Let $h(p_k,\xi_{ij})$ be the perpendicular
distance of the point $p_k$ to the straight line $\xi_{ij}$. 
The distance $l$ from a point $(x_0,y_0)$ to a straight line $ ax + by + c = 0$ 
is simply given by,
\begin{equation*}
l = \frac{|ax_0+by_0+c|}{\sqrt{a^2+b^2}}.
\end{equation*}

Let $d(p_i,p_j)$ be the Euclidean distance between points $p_i$ and $p_j$.
Let $L_{ij}$ be the geodesic length of the curve segment. 
Note that if the curve is not a straight line the Euclidean and geodesic distances
will be different. Then, we can define $f_{below}$,
$f_{upper}$, $f_{onTheLine}$ and $f_{bending}$ as features for
the points which are above the straight line, below the straight line,
on the straight line and reflect bending of the line, respectively. Mathematically:

\begin{equation}
f_{below} = max \Bigg(\frac{1}{N_a}  \sum_{p_k \in S_a} h(p_k,\xi_{ij}),
\frac{1}{N_b}  \sum_{p_k \in S_b} h(p_k,\xi_{ij})\Bigg),
\end{equation}

\begin{equation}
f_{upper} = min \Bigg(\frac{1}{N_a}  \sum_{p_k \in S_a} h(p_k,\xi_{ij}),
\frac{1}{N_b}  \sum_{p_k \in S_b} h(p_k,\xi_{ij})\Bigg),
\end{equation}

\begin{equation}
f_{bending} = \frac{L_{ij}}{d(p_i,p_j)}
\end{equation}

and

\begin{equation}
f_{onTheLine} = \{S_o\},
\end{equation}

\noindent where $N_a$ and $N_b$ are the number of points in $S_a$ and $S_b$
respectively. [Note that, the the points can also lie to the left and right
side of the line, depending on the orientation.]
We combine all the StringCut features as:
\begin{equation}
E_{StringCut} = f_{below} + f_{upper} + f_{onTheLine} + f_{bending}.
\end{equation}
After computing all the terms in the energy function in Equation (\ref{energy_eq}),
we have to optimize it, which is discussed in next section.
\end{enumerate}

\begin{figure}[htb!]
\begin{center}
\begin{tabular}{c c c}
\includegraphics[width=1.5in,height=1.0in]{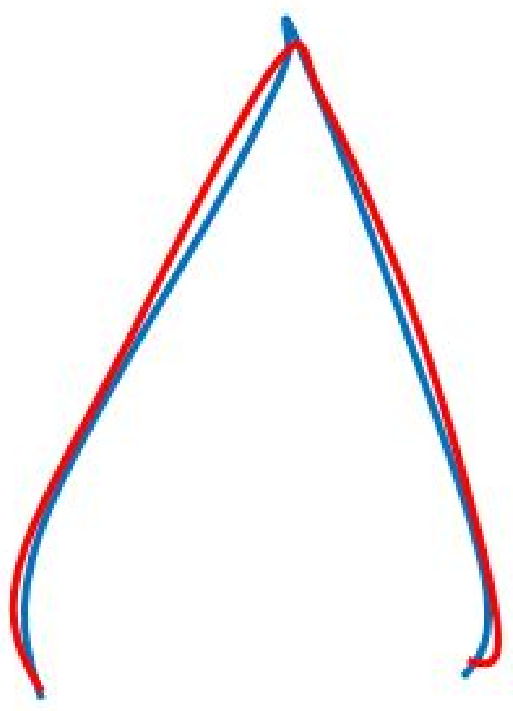} &
\includegraphics[width=1.5in,height=1.0in]{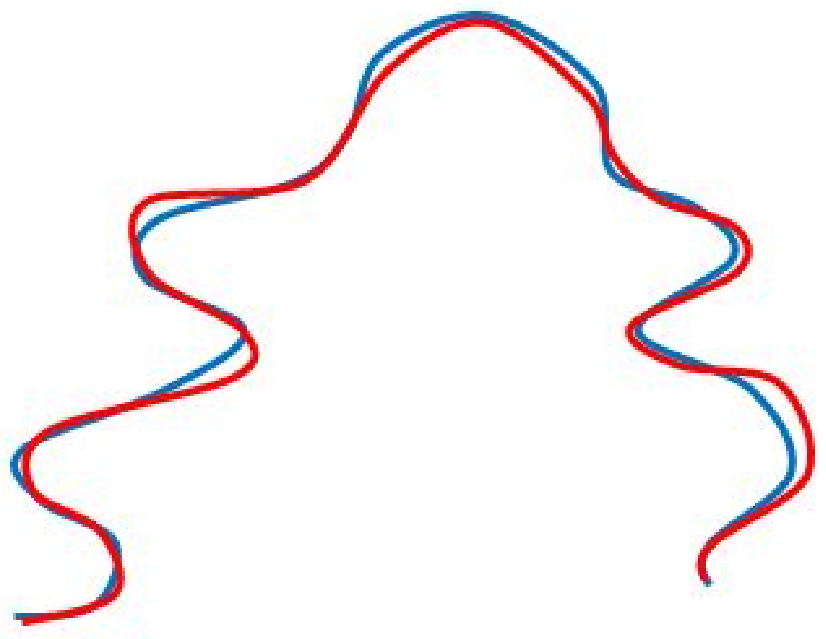} \\
(a) $F$=80,$E$=912 & (b) $F$=90,$E$=963 \\
\includegraphics[width=1.5in,height=1.0in]{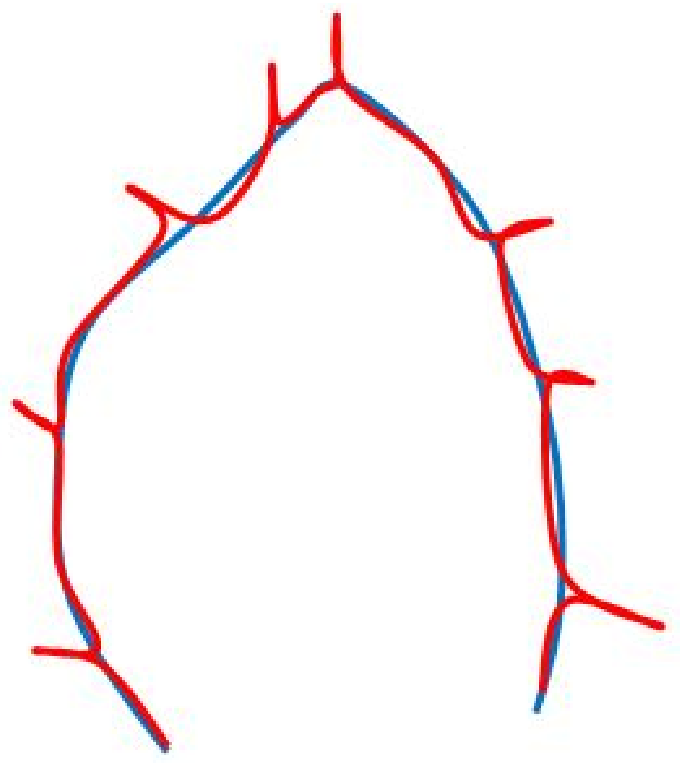} &
\includegraphics[width=1.5in,height=1.0in]{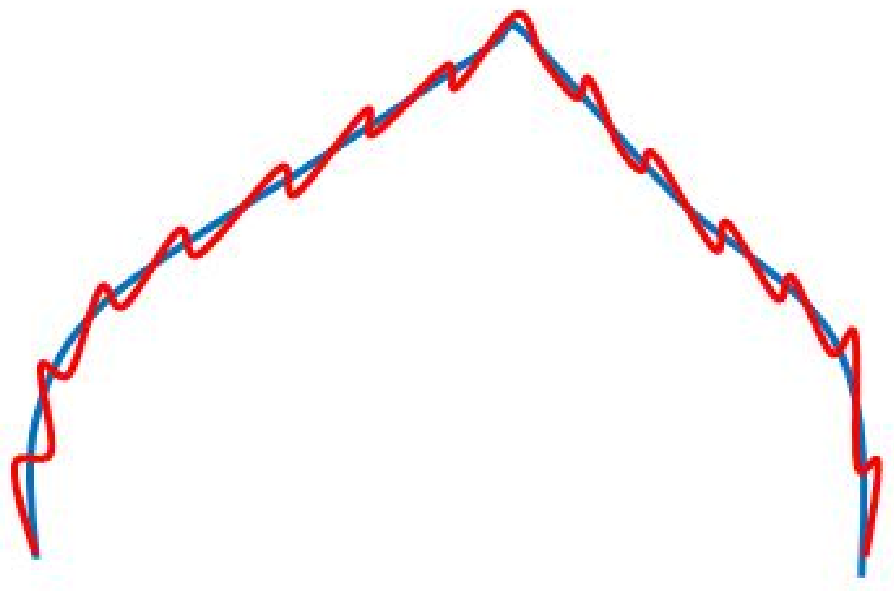} \\
(c) $F$=100,$E$=6322 & (d) $F$=91,$E$=7255\\
\includegraphics[width=1.5in,height=1.0in]{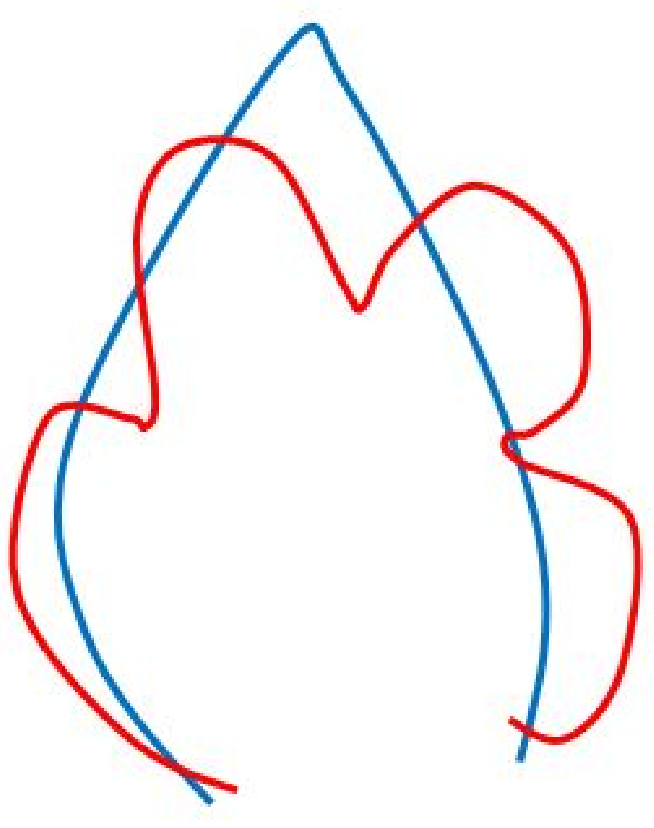} &
\includegraphics[width=1.5in,height=1.0in]{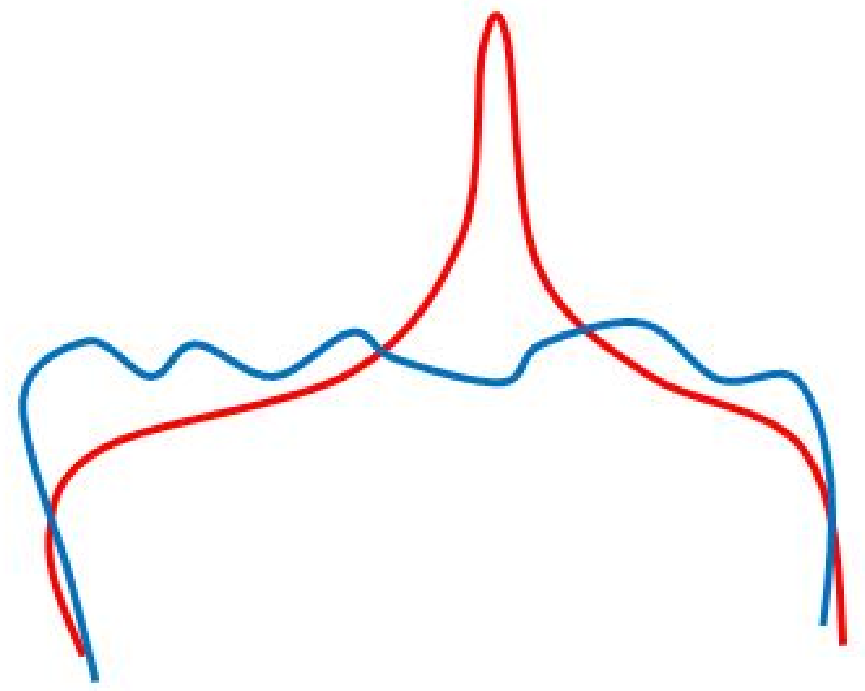} \\
(e) $F$=453,$E$=14432  & (f) $F$=523,$E$=16227 
\end{tabular}
\end{center}
\caption{Fr\'echet errors, $F$, and energy optimization values, $E$, for a number of curve sets.
Curves (a) and (b) have both low $F$ and $E$ values,
curves (c) and (d) have low $F$ and high $E$ values (showing the two are not always i
well correlated) and curves (e) and (f) have both high $F$ and $E$ values.}
\label{frechet_compare}
\end{figure}

It is instructive to consider the relationship between Fr\'echet matching error
and the energy optimization values. Again, this  demonstrates the need to use both.
We use the Fr\'echet distance metric to cheaply compute the similarity
of the two curves. The larger the curve differences,
generally the greater the Fr\'echet errors.
On the other hand, similar curves tend to have lower Fr\'echet errors.
Consider Figure \ref{frechet_compare}. Different curves are
shown in the blue and red colours, and are overlaid (as determined from the inverse affine
transform being applied to the results of the subgraph matching). The top row shows examples
of curves which are similar in terms of global structure and local structures.
In this case, low Fr\'echet errors, $F$, are well correlated with low optimization
energy, $E$. The bottom row shows examples where the curves are very different. 
Now the $F$ and $E$ values are again well correlated but now are much larger, 
indicating poor matches.
The middle row shows some (rare) cases, where
the Fr\'echet errors are low but the energy values are high. Fr\'echet thresholding alone
would fail to remove these bad matches in this case, but expensive 
energy optimization calculation would remove the matches. 
The Fr\'echet error, by itself, would correctly
accept/reject the matches in the top and bottom rows.
Thus, the Fr\'echet errors and energy optimization play a complementary role of reduced
computational costs with reasonable accuracy versus high accuracy at considerably increased 
computational costs.

In the course of our experimentation we noticed that the energy functional
was necessary 20\% to 40\% of the time to determine the best overall match. 
These cases occurred when there were
significant local variations in the leaves; in these cases the lowest Fr\'echet error
did not correspond to the best match (but one of the other $\tau$-1 matches did).
Again $\tau$ was empirically determined as 5 for the datasets used in this paper.

\section{GNCCP optimization}

The energy functional defined in Equation (\ref{energy_eq}) can be optimized in different ways. 
Because of the need for computational efficiency, we adopted a recently proposed efficient
Graduated Non-Convex and Concavity Procedure (GNCCP)
\cite{GNCCP-PAMI-2014} approach. 
GNCCP basically solves the combinatorial optimization problem
using permutation matrices. In our case, we have formulated the energy
functional as a weighted combination of adjacency matrices. In another words, the problem
is formulated as an assignment problem, where we explore all possible combinations
of discrete points on the curve to find the best match (note that we are dealing
with open curves at this stage). The adjacency matrix automatically handles the
ordering of points (we chose the counter-clockwise direction). Also, all the curve
sections have the same number of points, which is obtained by representing the
original points by a $\beta$-spline and re-sampling the spline
by controlling the increment parameter $\Delta u$, to
obtain any specified number of equally spaced spline points.

The energy functional in our case is a combination of adjacency matrices,
whose optimization gives the one-to-one correspondence of the points. 
Given two graphs, $G_1=(E_1,V_1)$ and $G_2=(E_2,V_2)$, where $E$ and $V$ are the set of
edges and vertices respectively, a one-to-one mapping function 
$f:V_1 \rightarrow V_2$ specifies the correspondence between the two graphs. A typical
approach to finding this solution is to formulate an energy functional, $f$, and minimize it.
In general, this is an NP-hard combinatorial optimization problem
\cite{GNCCP-PAMI-2014}. 

Let the two graphs have $M$ and $N$ vertices respectively. We want to find the
optimal matching of vertices between the two graphs. Let $c_{ij}$ denote the
cost of matching the $i$-th vertex of $G_1$ with the $j$-th  vertex of $G_2$.
Let $x_{ij} \in \{0,1\}$ denotes the assignment. Then, the optimization 
problem can be written as:
\begin{equation}
\begin{split}
\min_{\mathbf{X} \in \mathcal{P}} F(\mathbf{X}) = \arg\min_x \sum_{i=1}^{M} \sum_{j=1}^{N} c_{ij} x_{ij} \\
=  \arg\min_\mathbf{X} \mathbf{X}^T \mathbf{C},
\end{split}
\end{equation}
such that the following  conditions are satisfied:
\begin{equation}
\begin{split}
\mathbf{X} \in \mathcal{P}, \mathcal{P} := 
\Big\{\mathbf{X}|x_{ij} = \{0,1\}, \\
\sum_{j=1}^{N} x_{ij} = 1, \sum_{i=1}^{M} x_{ij} \leq 1, \forall i,j \Big\}, M \leq N,
\end{split}
\label{gnccp_formulation}
\end{equation}

\noindent where  $\mathcal{P}$ is the set of $(M \times N)$ permutation matrices.

%

Following the approach by Maciel \emph{et al.} \cite{sparseCorr-PAMI-2003},
the domain of the problem is relaxed from $\mathcal{P}$ to its convex hull, which is
the set of doubly stochastic matrices, $\Omega$, given as follows:

\begin{equation}
\Omega := 
\Big\{\mathbf{X}|x_{ij} \geq 0, \sum_{j=1}^{N} x_{ij} = 1, \sum_{i=1}^{M} x_{ij} \leq 1, \forall i,j\Big\}.
\end{equation}

Then the GNCCP algorithm approximately solves the above problem as,

\begin{equation}
F_{\zeta}(\mathbf{X}) = 
\begin{cases}
(1-\zeta) F(\mathbf{X}) + \zeta tr(\mathbf{X}^T \mathbf{X}) \hspace{10mm} if \hspace{2mm} 1 \geq \zeta \geq 0 \\
\\
(1+\zeta) F(\mathbf{X}) + \zeta tr(\mathbf{X}^T \mathbf{X}) \hspace{8mm} if \hspace{2mm} 0 \geq \zeta \ge -1,
\end{cases}
\end{equation}

\noindent where  $\mathbf{X} \in \Omega$ and and $tr(\cdot)$ denotes the 
trace of a matrix. As the algorithm converges, 
the variable $\zeta$ decreases from 1 to -1. The steps are shown in
Algorithm \ref{gnccp}.

\begin{algorithm}
\caption{GNCCP algorithm}\label{gnccp}
\begin{algorithmic}[1]
\State $\zeta \gets 1$, $\mathbf{X} \gets \mathbf{X}^0$
\While{$\zeta > -1 \land \mathbf{X} \notin \mathcal{P}$ }
\While{$\mathbf{X}$ has not converged}
\State $\mathbf{Y} = \arg\min_{Y} tr(\nabla F_{\zeta}(\mathbf{X})^T \mathbf{Y})$,  $\mathbf{Y} \in \Omega$
\State $\alpha = \arg\min_{\alpha} F_{\zeta}(\mathbf{X}+\alpha(\mathbf{Y}-\mathbf{X}))$, $0 \leq \alpha \leq 1$
\State $\mathbf{X} \gets \mathbf{X} + \alpha (\mathbf{Y} - \mathbf{X})$
\EndWhile
\State $\zeta \gets \zeta - d\zeta$
\EndWhile
\State \textbf{return} $\mathbf{X}$
\end{algorithmic}
\end{algorithm}

From the energy optimization performed as discussed above, we rank the
$\eta$ best Fr\'echet curves according to their
energy values and choose as the best match the curve having the minimum energy. 

\section{Experimental Results}

We validate the proposed method by applying it to $3$ publicly available leaf
datasets: the Swedish dataset \cite{swedishleafThesis-2001},
the Flavia dataset \cite{flavia-2007} and
the Leafsnap dataset \cite{leafsnap-ECCV-2012}.
As pointed out by Hu \emph{et al.} \cite{MDM-TIP-2012},
the Smithsonian leaf dataset \cite{IDSC-PAMI-2007}
contains too few samples per species, which makes it unsuitable for
extensive experimentation.
To make fair comparisons with the other algorithms, we compare our method
with state-of-the-art algorithms for which code is publicly available, or whose
implementation is straightforward (the paper contains enough details to
implement the idea). We compare our algorithm with the Shape Context (SC)\footnote{\url{https://www2.eecs.berkeley.edu/Research/Projects/CS/vision/shape/sc_digits.html}}
\cite{SC-PAMI-2002}
and the Inner Distance (IDSC)\footnote{\url{http://www.dabi.temple.edu/~hbling/code_data.htm}}
\cite{IDSC-PAMI-2007} methods. Implementations of both
algorithms are publicly available. IDSC actually uses SC but improves it
by using Dynamic Programming (DP).
Also we have implemented the Multiscale Distance Matrix (MDM) method \cite{MDM-TIP-2012}.

\begin{figure}[htb!]
\begin{center}
\begin{tabular}{c c}
\includegraphics[width=1.5in,height=1.0in]{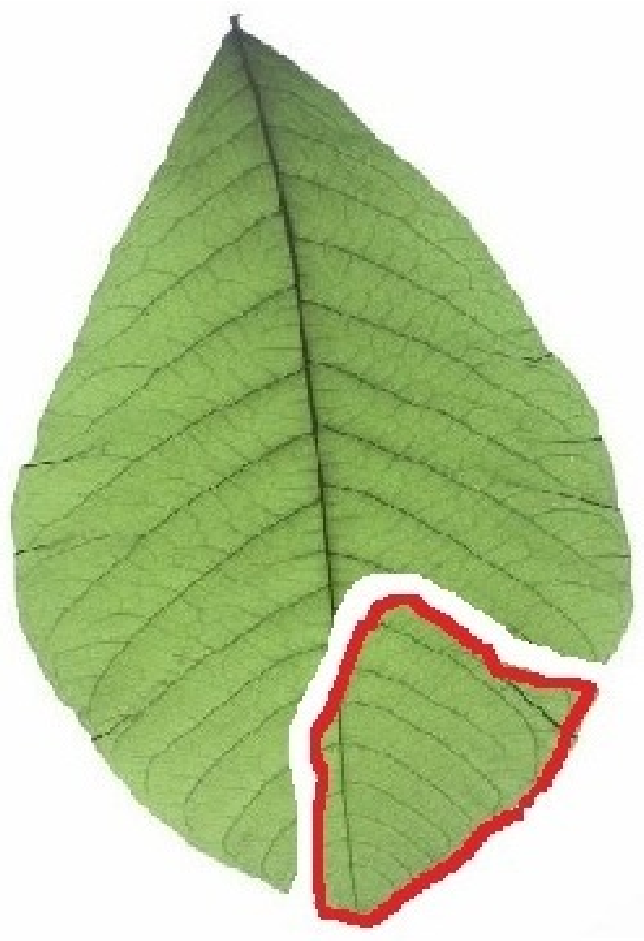} &
\includegraphics[width=1.5in,height=1.0in]{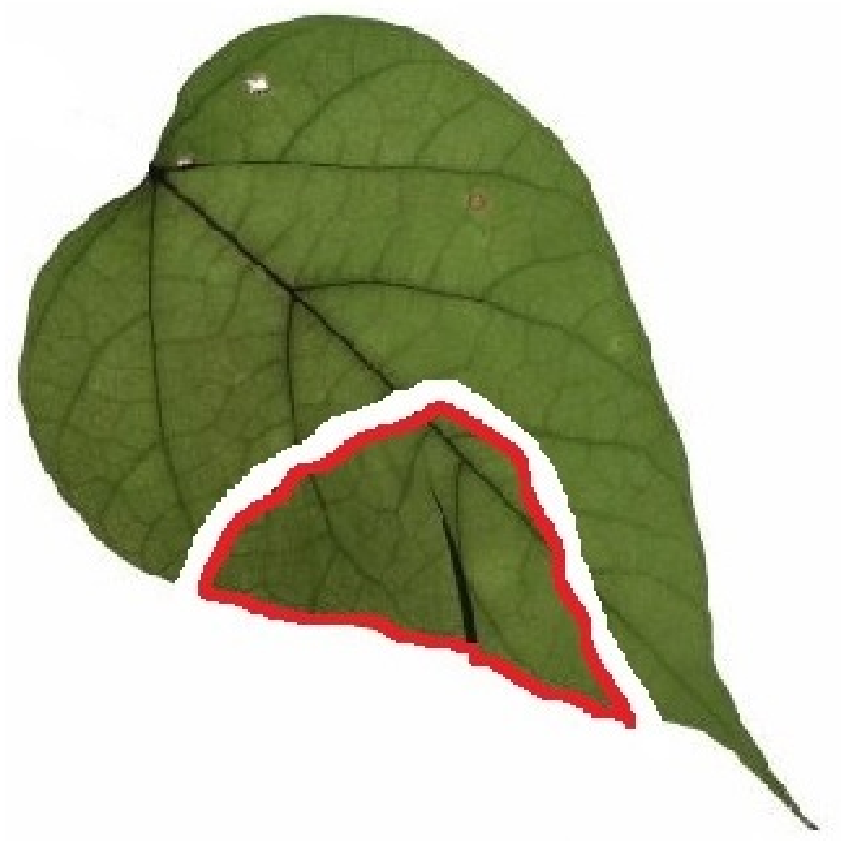} \\
(a) 18\% & (b) 21\%\\
\includegraphics[width=1.5in,height=1.0in]{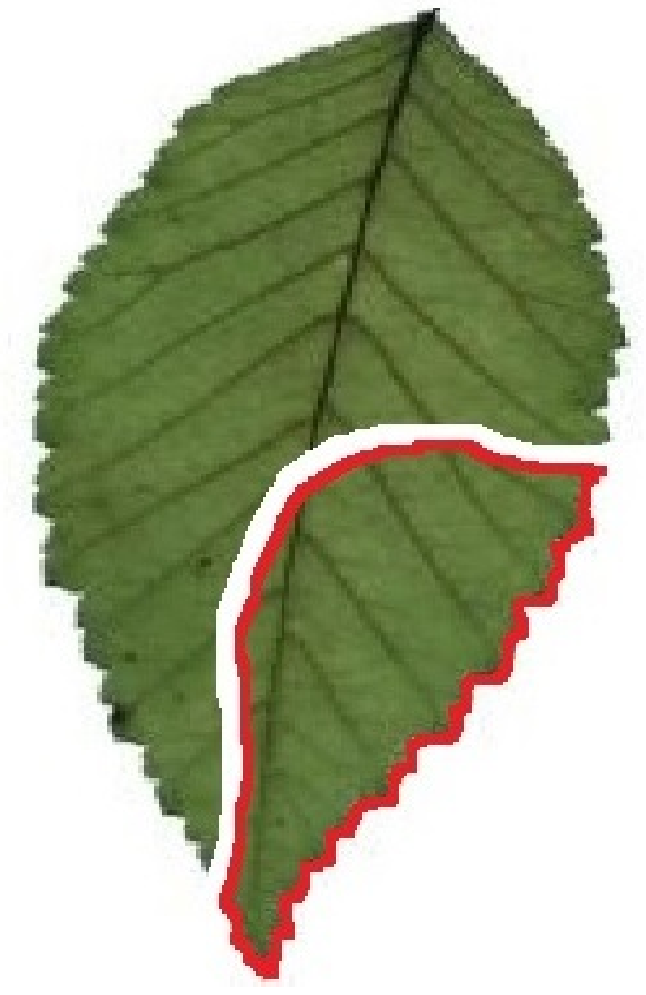} &
\includegraphics[width=1.5in,height=1.0in]{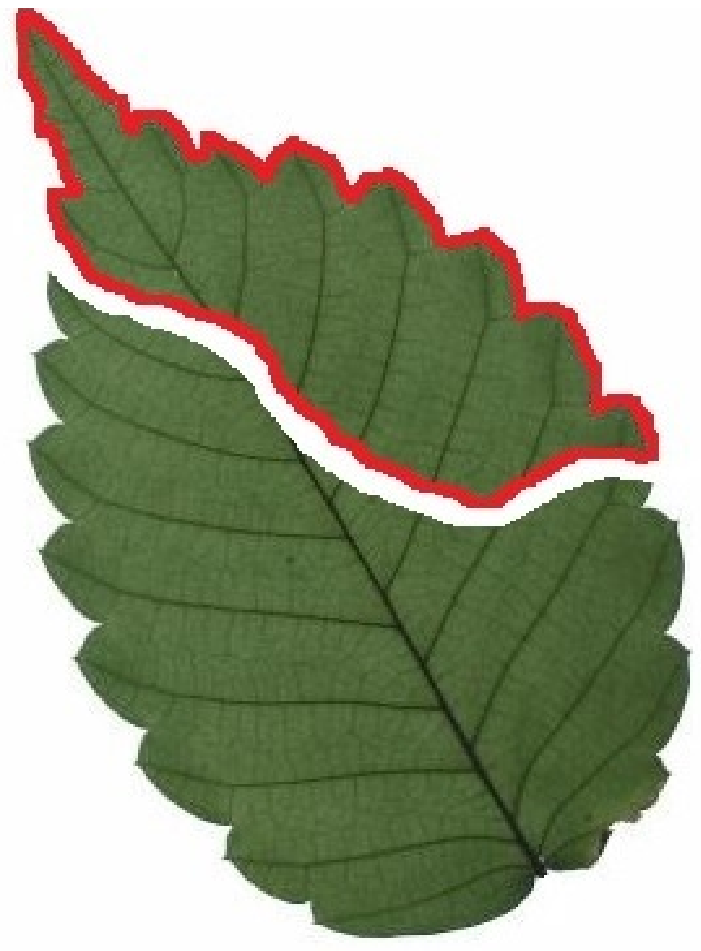} \\
(c) 29\% & (d) 36\% \\
\includegraphics[width=1.5in,height=1.0in]{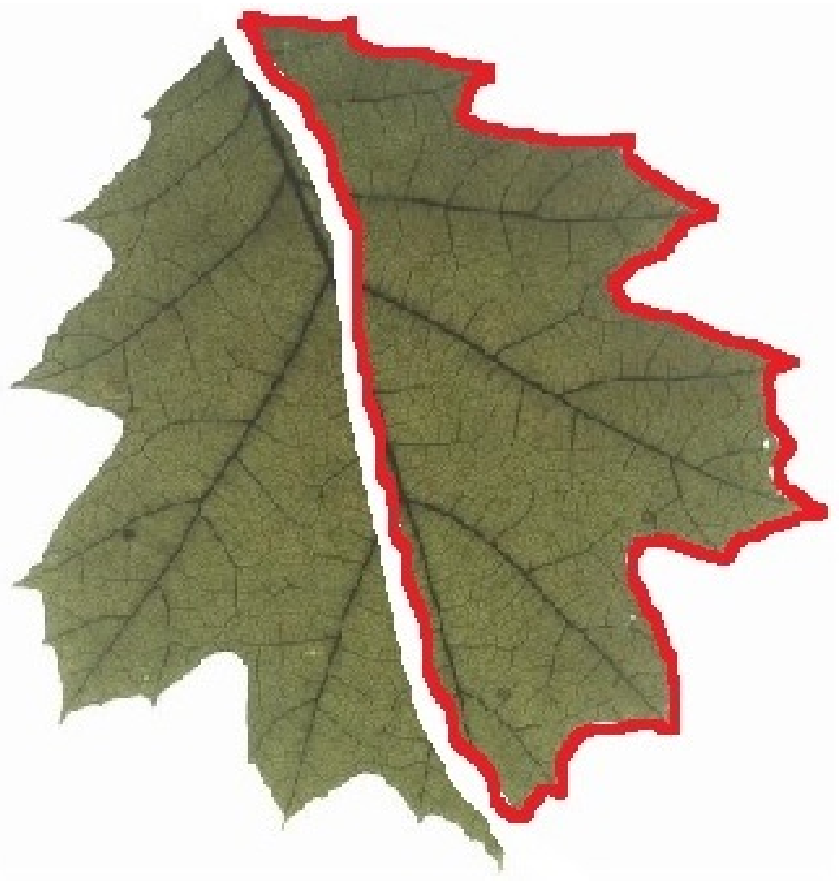} &
\includegraphics[width=1.5in,height=1.0in]{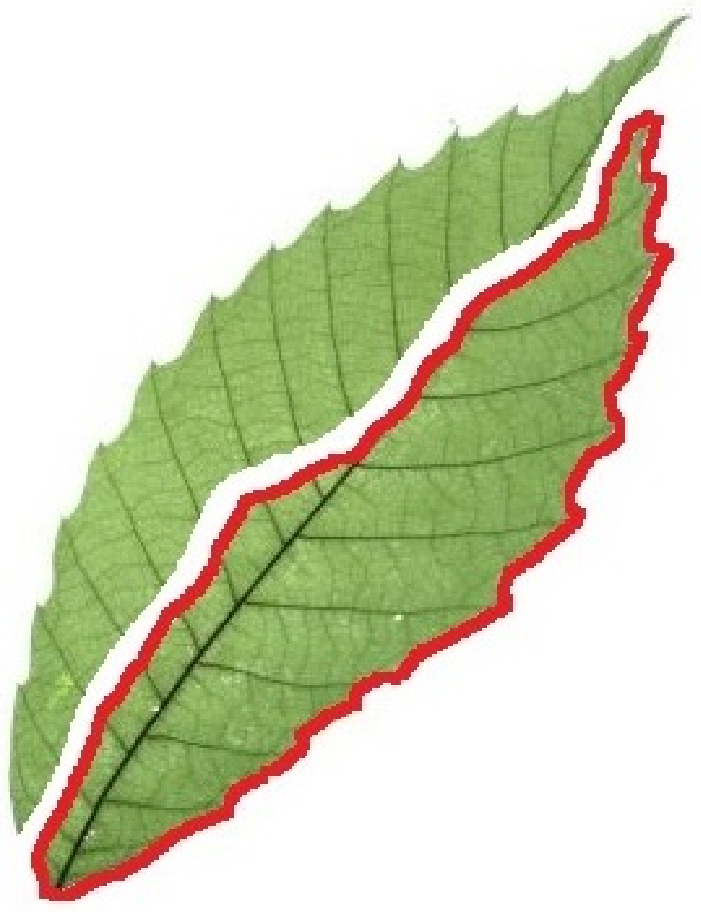} \\
(e) 50\% & (f) 49\%
\end{tabular}
\end{center}
\caption{Some typical occluded leaves with varying contour occlusion levels (specified
below each image as a percentage). The part of each leaf enclosed by a red contour
is cut away. The partial leaf contour remaining (not including the contour the cut away
leaf part makes with the leaf) is the open occluded curve.}
\label{occlusion_examples}
\end{figure}

\begin{figure}[htb!]
\begin{center}
\begin{tabular}{c c}
\includegraphics[width=1.5in,height=0.95in]{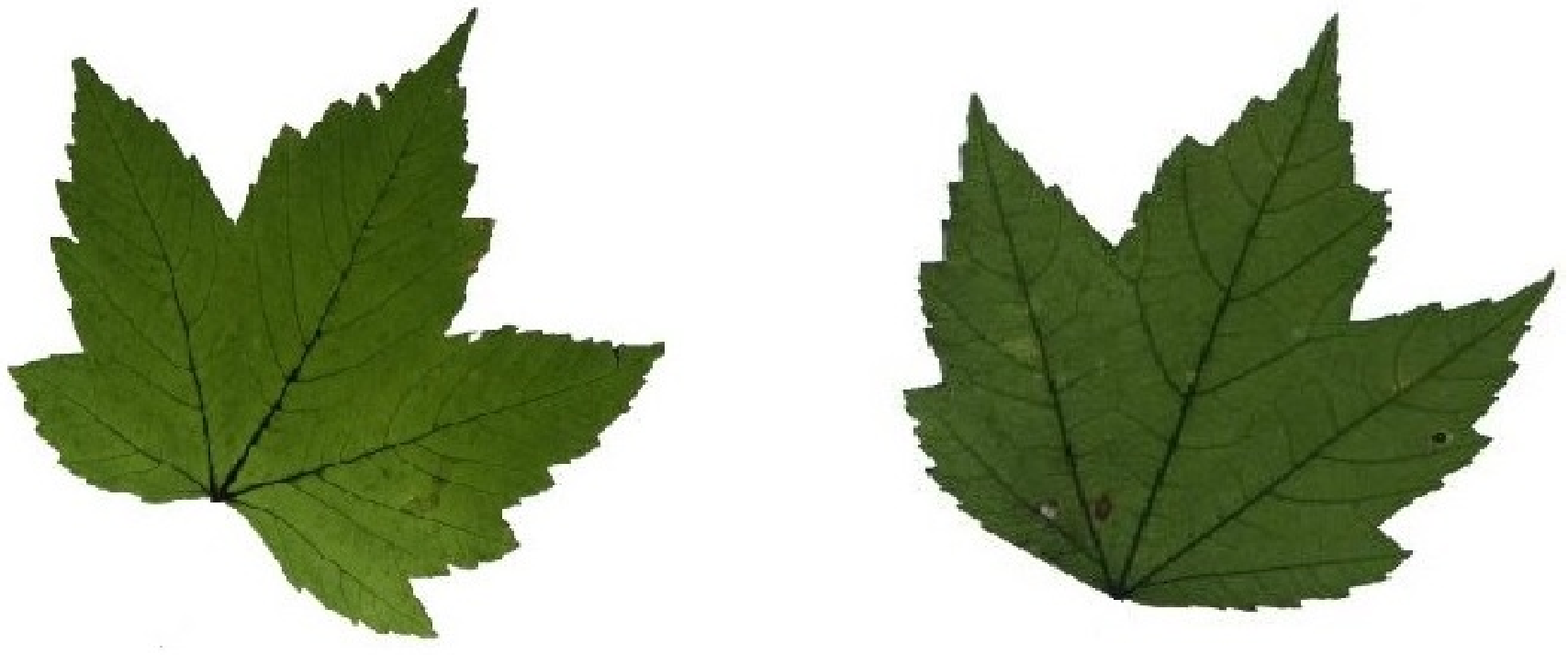} &
\includegraphics[width=1.5in,height=0.95in]{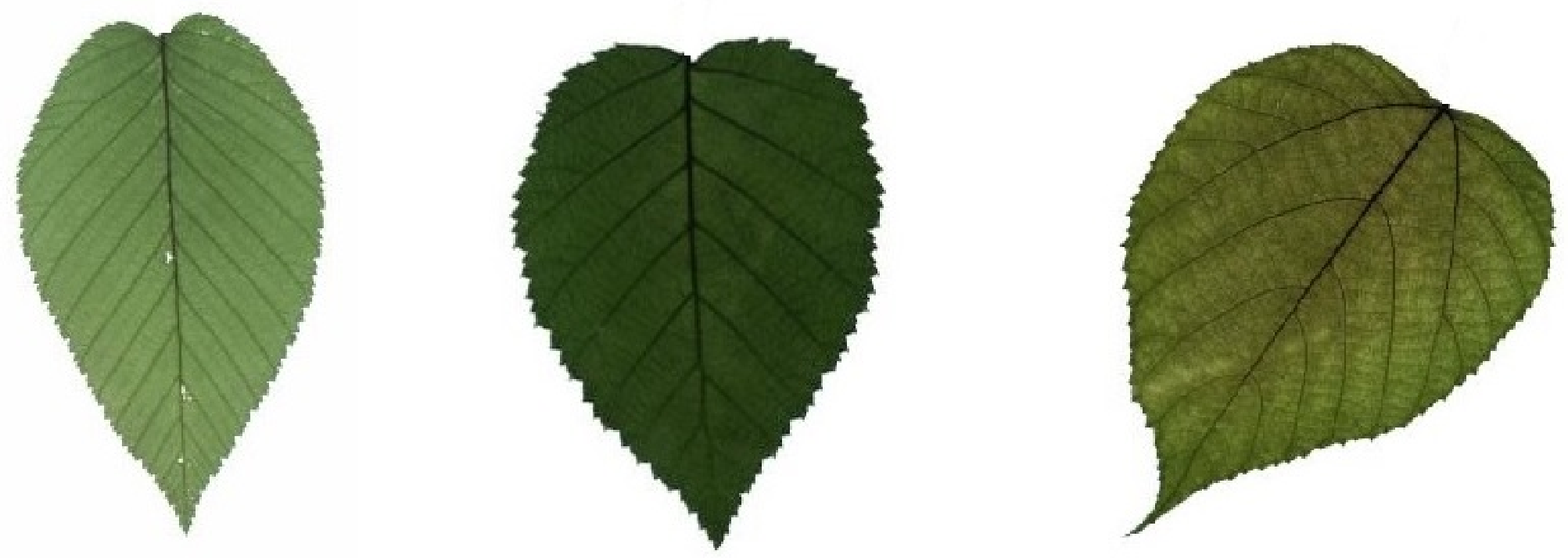} \\
(a) & (b) \\
\includegraphics[width=1.5in,height=0.95in]{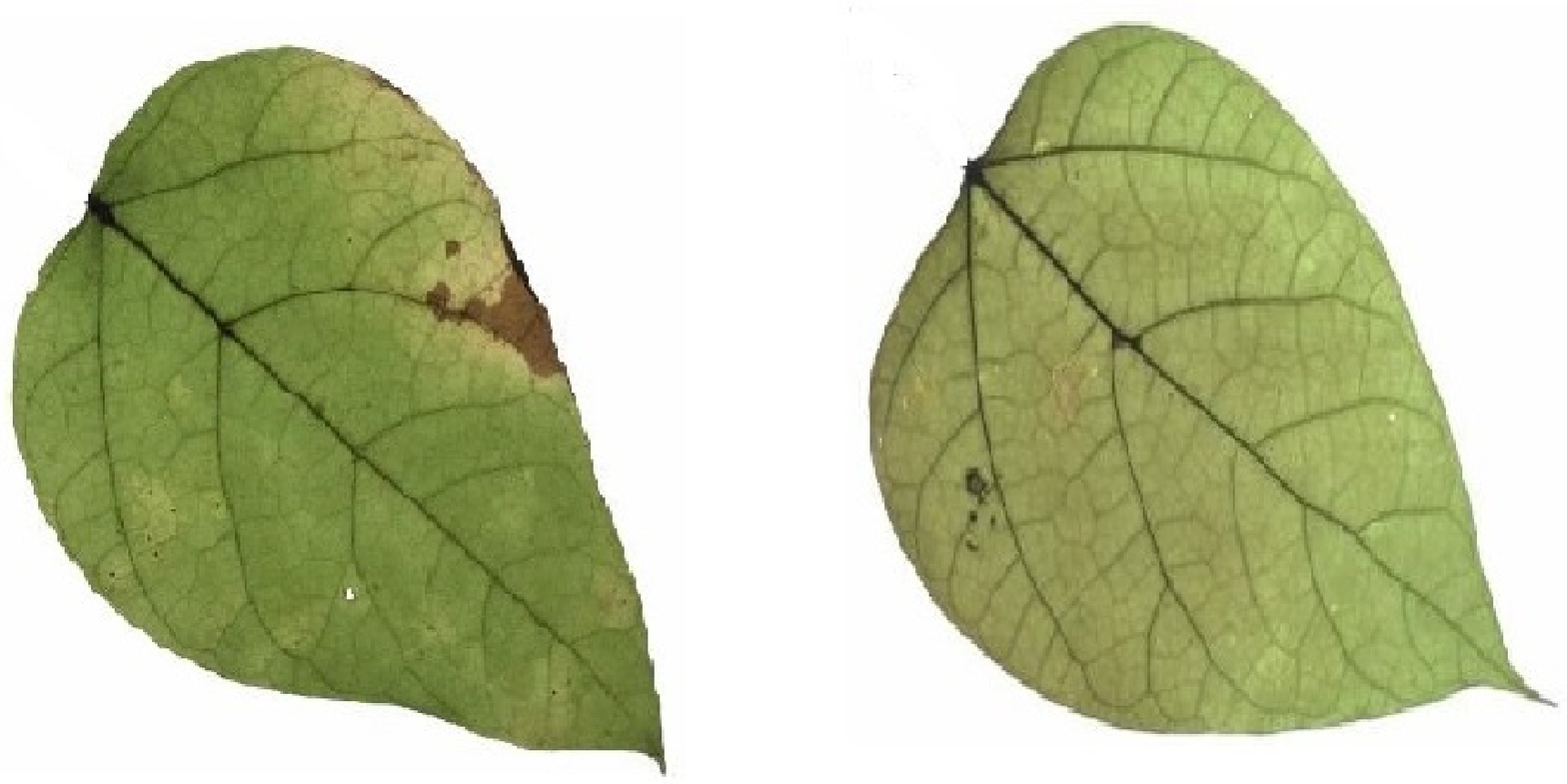} &
\includegraphics[width=1.5in,height=0.95in]{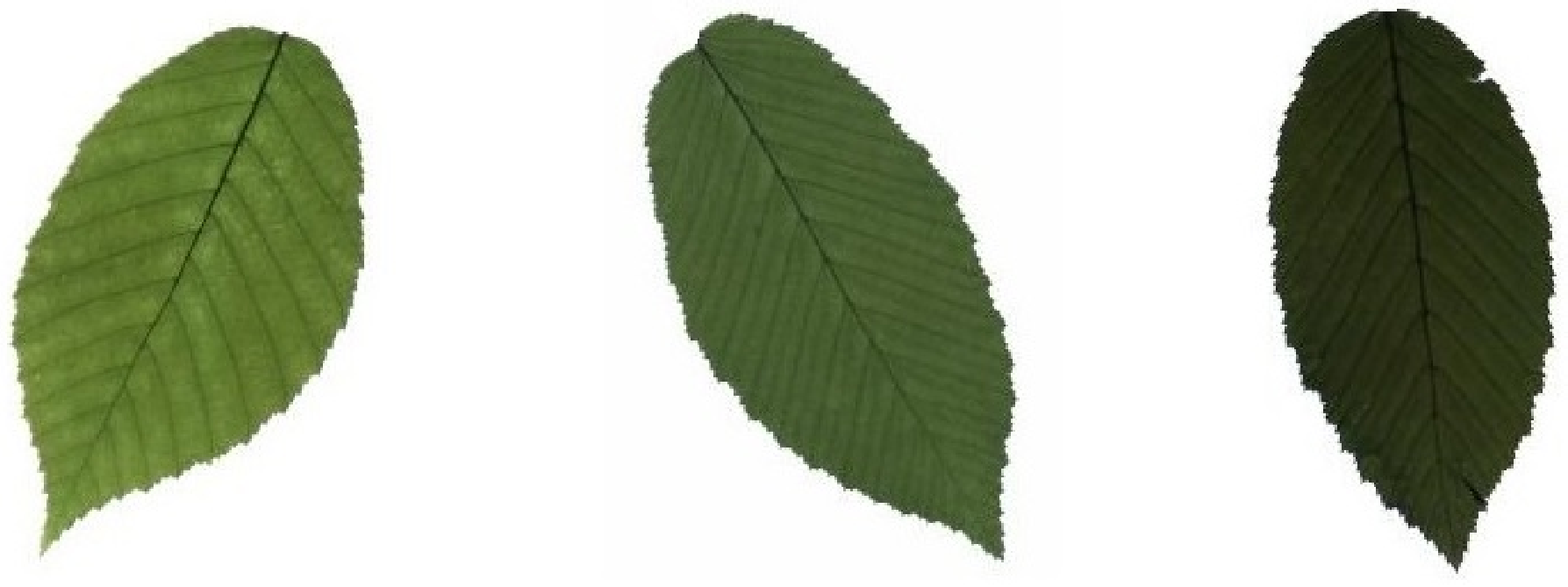} \\
(c) & (d) \\
\end{tabular}
\end{center}
\vspace*{-0.1in}
\caption{Leaves with similar contours for different species: leaves from
(a) Sycamore Maple (Acer Pseudoplatanus) and Red Maple (Acer Rubrum),
(b) Sweet Birch (Betula Lenta), Whitebark Himalayan Birch (Betula Jacquemontii) and
Paper Mulberry (Broussonetxtia Papyrifera),
(c) Southern Catalpa (Catalpa Bignonioide) and Northern Catalpa (Catalpa Speciosa) and
(d) Yellow Birch (Betula Alleghaniensis), Downy Serviceberry (Amelanchier Arborea) and
European Hornbeam (Carpinus Betulus). Note that we only classify these leaves based on their
contour shapes and not on their interior textures, even though some of the species are
clearly distinguishable by their texture differences.}
\label{similar_leaves}
\end{figure}


For occlusion, we randomly apply $20\%$ to $50\%$ occlusion to the test
leaf. By this we mean the percentage of the overall contour that is occluded.
We do not consider how much leaf area is occluded.
We also assume that the occlusion boundary is known. That is, we have an
open curve as input. Occlusion boundary detection is a different problem, and that
is not the focus of the paper.\footnote{Another research project we are engaged with involves 
range sensing a plant over time to measure its growth. The sensor we are using allows
registered depth and grayvalue images of specified plant leaves, allowing the use of depth and 
intensity discontinuities to segment individual leaves from the matter around them. 
We then want the capability to track 
individual leaves over time as they grow and potentially become occluded by other plant parts.
The work described in this paper will be very useful for performing this task.}
Figure \ref{occlusion_examples} shows some examples of occluded leaves.
The top row shows lower levels of occlusion, the middle row shows medium levels of
occlusion, and the last row shows examples when the leaves are highly 
occluded (up to $50\%$). Note that the occlusion level is considered at the
contour, occlusion in the interior of the leaves are not taken into consideration.
A leaf can be $90\%$ occluded in terms of it's area, and the occlusion at
the contour can be $10\%$! Our algorithm may be able to handle these cases since we classify
the leaves by their contours. We do not match leaves based on their interior 
textures in this paper.

Sometimes, it is difficult to distinguish the leaf species if the amount of occlusion is
high. For example, consider Figure \ref{similar_leaves}, which shows 4 leaf sets with
leaves from different species. Even for small occlusion levels, discriminating among the
leaf species is difficult. At $50\%$ occlusion it would be impossible.
In our work, we consider any of close leaf species as the correct
answer to the classifier. Classifying these types of occlusion cases
is out of the scope of this paper (the same issue arises for full leaf matching
and was handled in the same way \cite{IDSC-PAMI-2007}).
We present and discuss different datasets and 
corresponding recognition results below.

\subsection{Swedish dataset}
\begin{figure*}
\begin{center}
\begin{tabular}{c c}
\includegraphics[width=.48\textwidth]{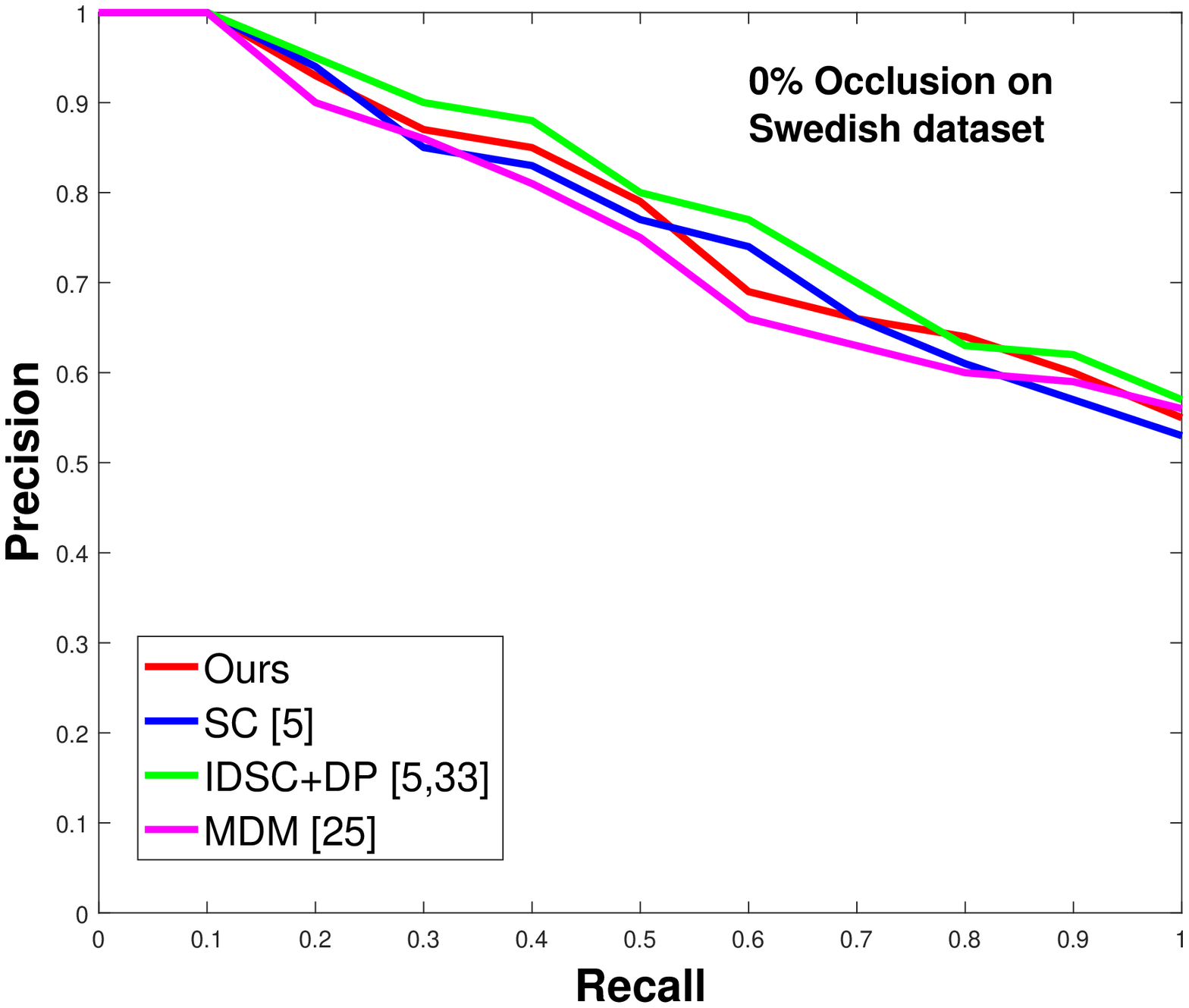} &
\includegraphics[width=.48\textwidth]{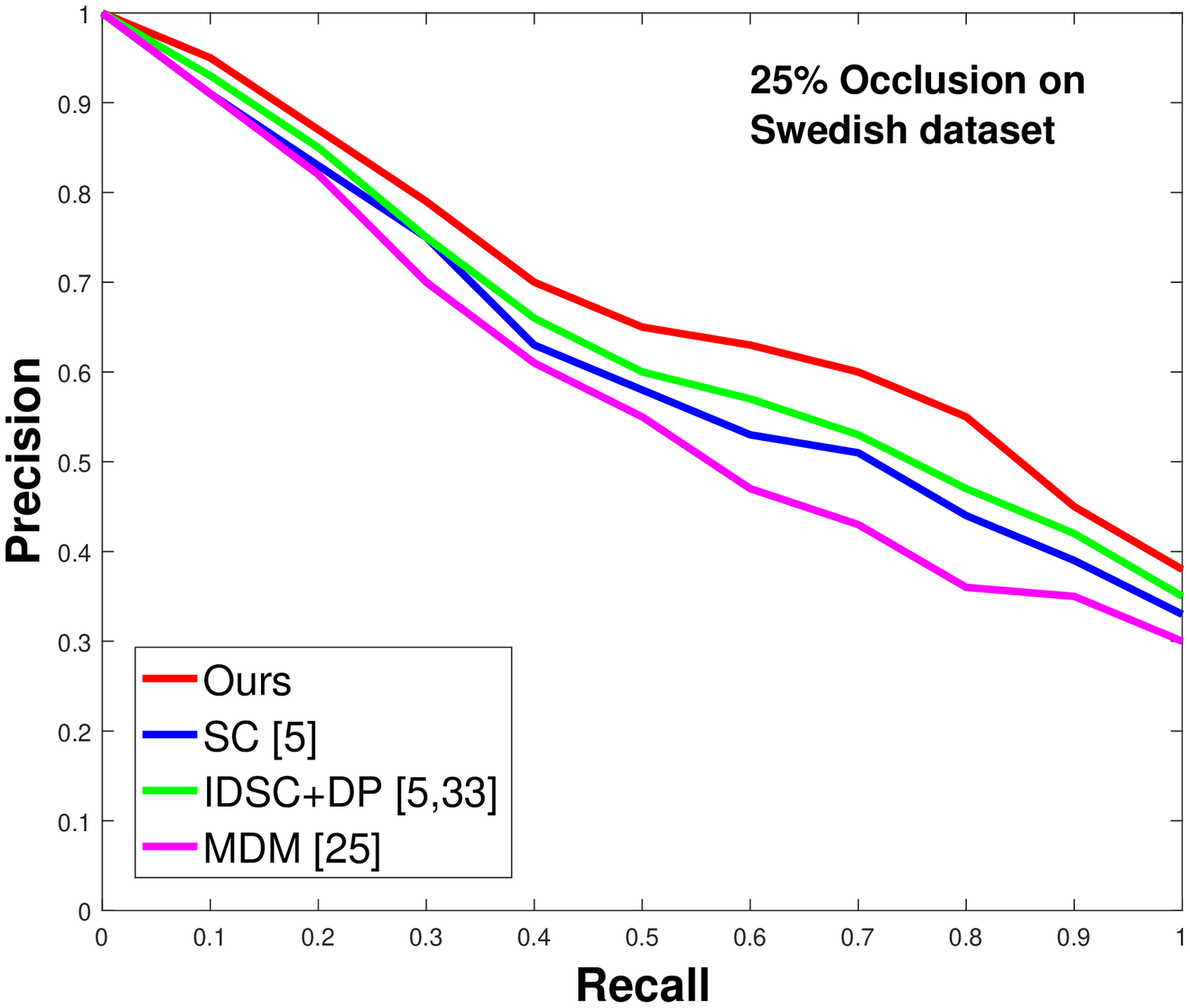} \\
(a) & (b) \\
\includegraphics[width=.48\textwidth]{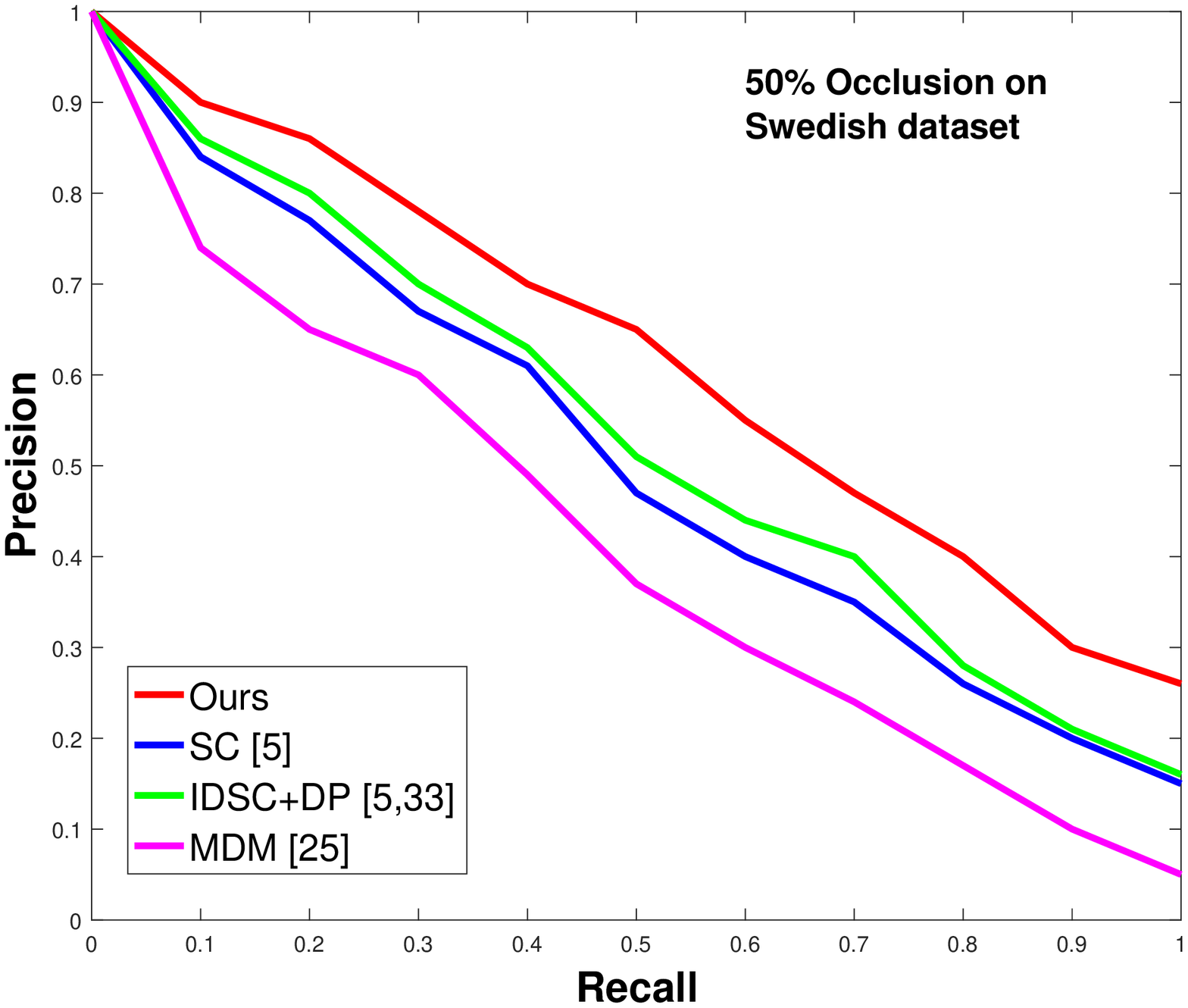} &
\includegraphics[width=.48\textwidth]{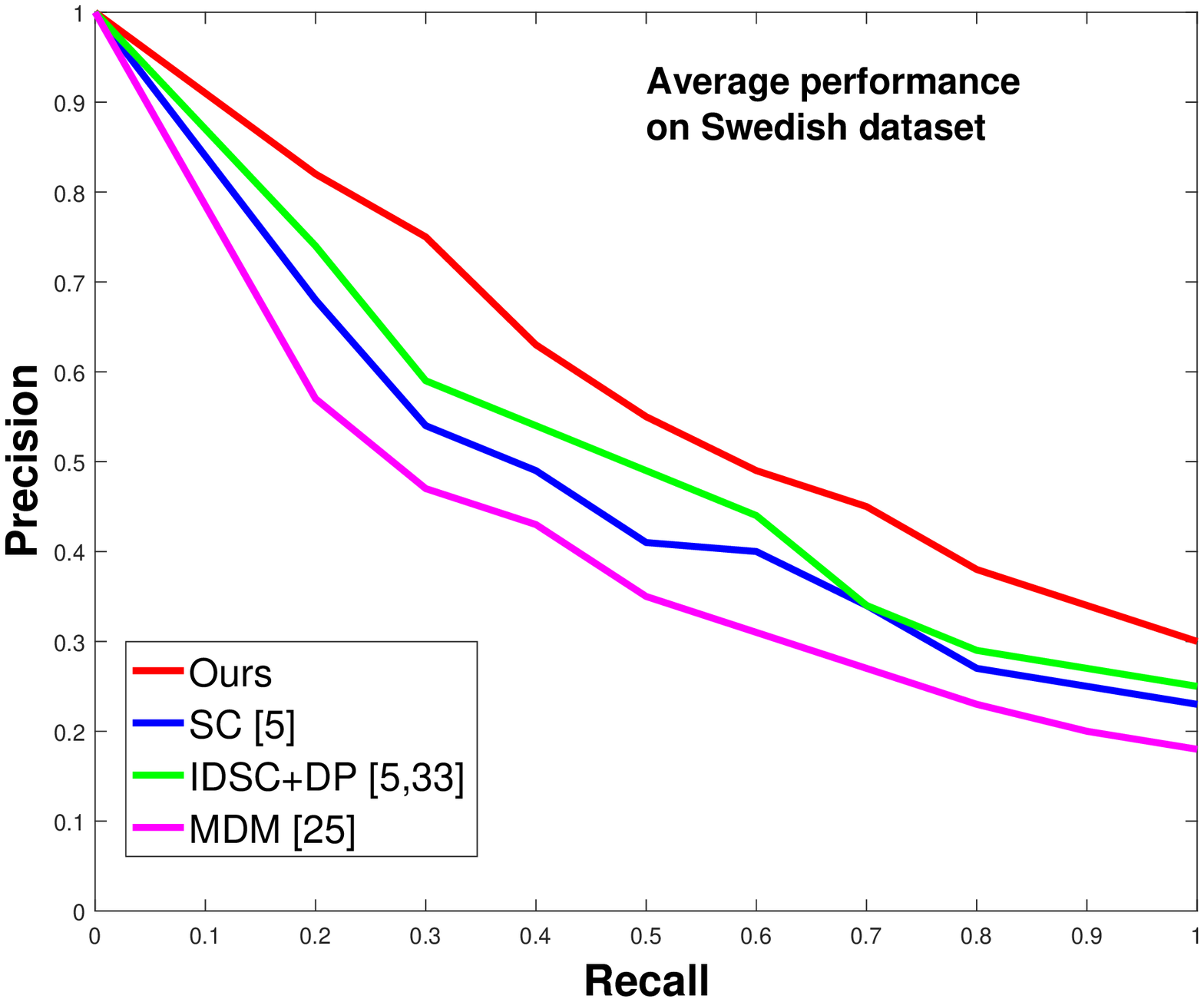}   \\
(c) & (d)
\end{tabular}
\end{center}
\caption{Precision-Recall curves for the Swedish dataset with (a) no occlusion, 
(b) $25\%$ occlusion, (c) $50\%$ occlusion and (d) average performance for random 
occlusion levels from $20\%$ to $50\%$ respectively.}
\label{swedish_pr}
\end{figure*}

This dataset contains $1125$ leaves from $15$ species with $75$ images per species.
Following the protocol of previous work (\cite{IDSC-PAMI-2007}, \cite{shapeTree-CVPR-2007},
\cite{MDM-TIP-2012}, \cite{stringCut-TIP-2014}, \cite{leafCounting-PR-2015}),
we chose $25$ images as database images and the rest as the testing images for each 
species. Also, we have removed the stems from the leaves \cite{MDM-TIP-2012}. 
We applied 0\%, 25\% and 50\% random occlusion to the leaves.
As stated previously, we also applied random occlusion to the contour, ranging
from $20\%$ to $50\%$. We present the results of $0\%$ occlusion (full leaf),
$25\%$ occlusion, $50\%$ occlusion and the average performance of the algorithm for
random occlusion levels from $20\%$ to $50\%$
as precision-recall curves in Figure \ref{swedish_pr}.
Figure \ref{swedish_pr}a shows the performance when there is no occlusion. 
The performance of
our algorithm is similar to the state-of-the-art. However, as the occlusion
level is increased, our algorithm starts outperforming the state-of-the-art. As shown
in Figure \ref{swedish_pr}d, the average
performance of our algorithm is significantly better than the state of the art for
higher amounts of occlusion.

\subsection{Flavia dataset}
\begin{figure*}
\begin{center}
\begin{tabular}{c c}
\includegraphics[width=.48\textwidth]{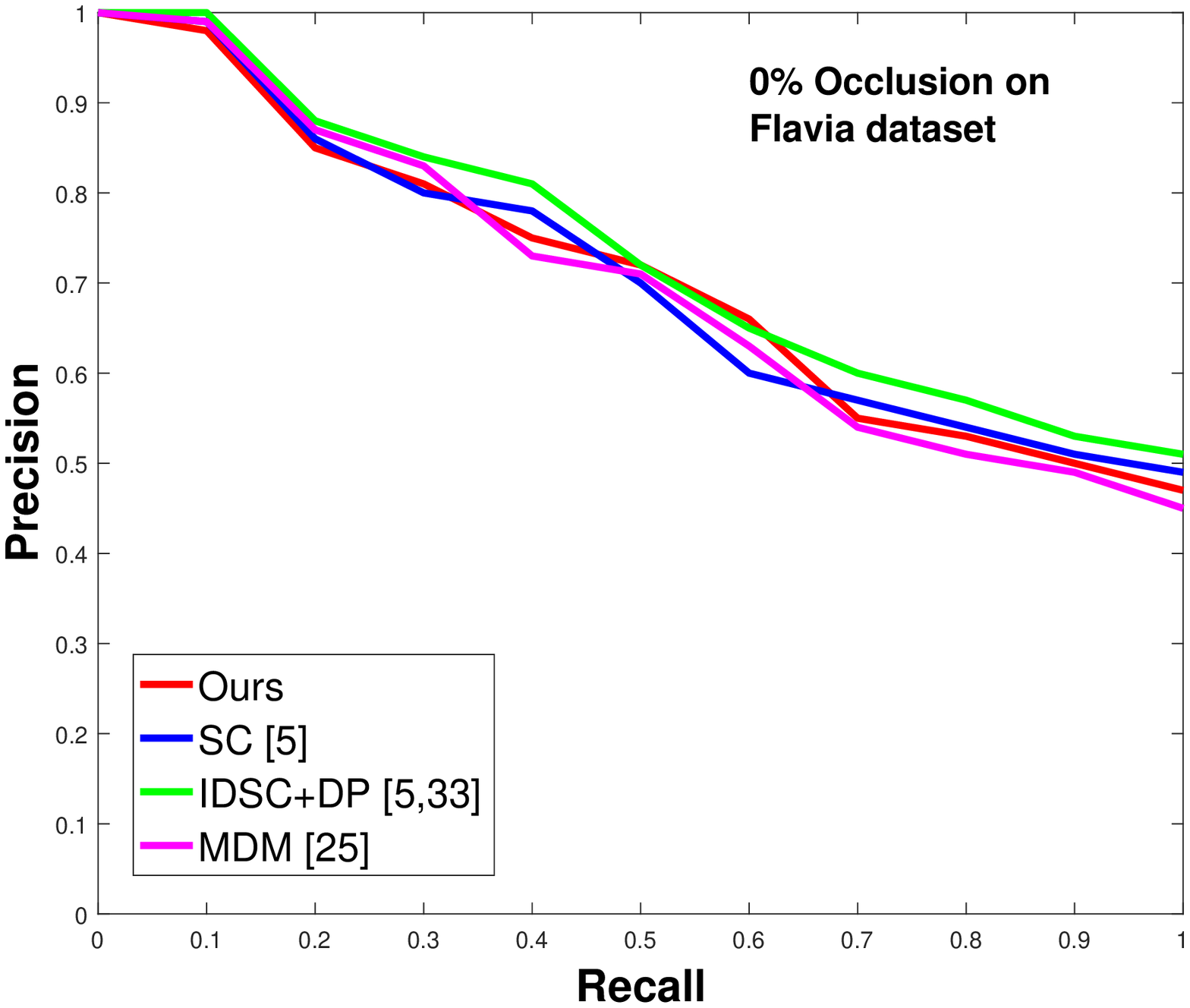} &
\includegraphics[width=.48\textwidth]{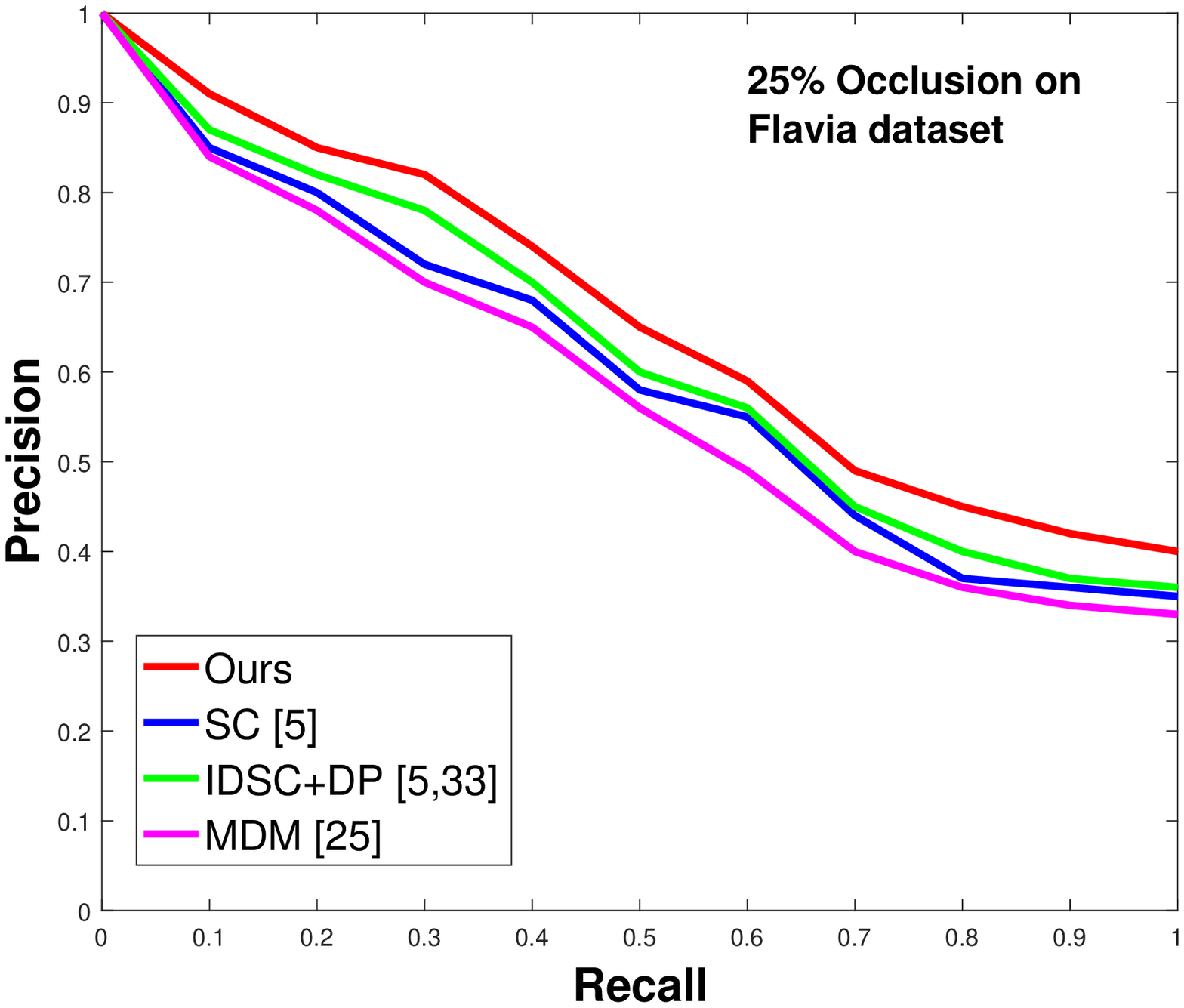} \\
(a) & (b) \\
\includegraphics[width=.48\textwidth]{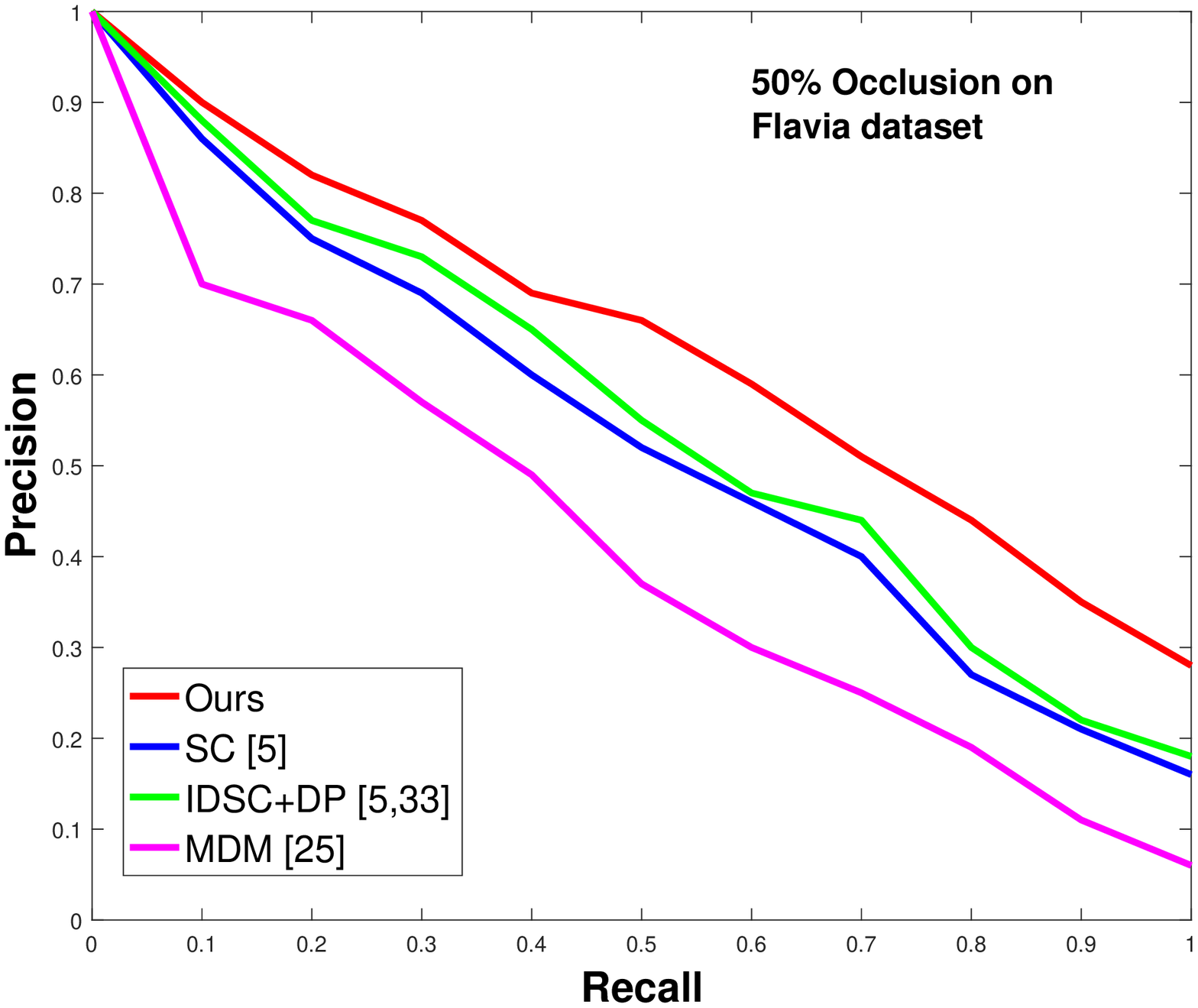} &
\includegraphics[width=.48\textwidth]{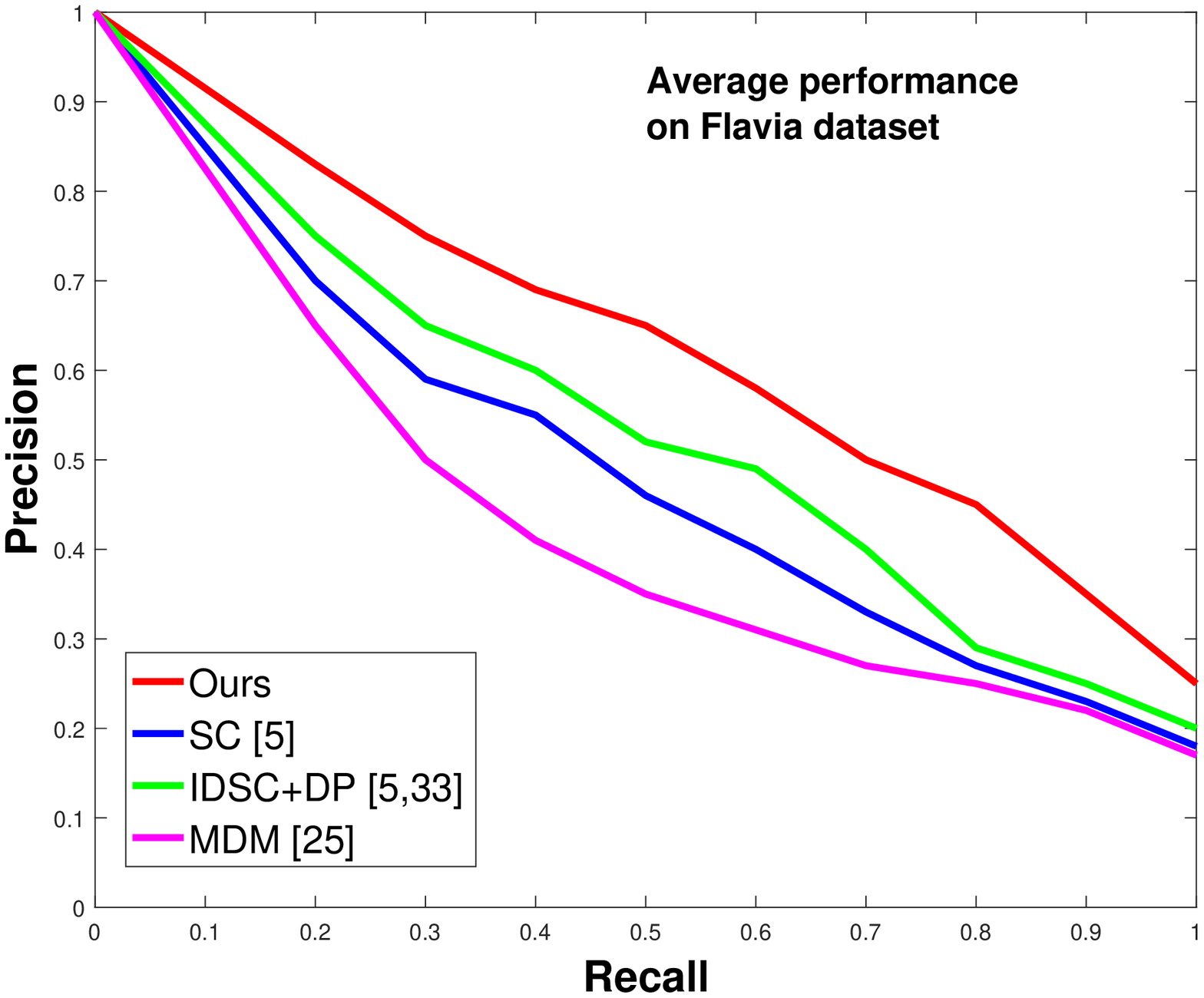}   \\
(c) & (d)
\end{tabular}
\end{center}
\caption{Precision-Recall curves for the Flavia dataset with (a) no occlusion, 
(b) $25\%$ occlusion, (c) $50\%$ occlusion and (d) average performance for random 
occlusion levels from $20\%$ to $50\%$ respectively.}
\label{flavia_pr}
\end{figure*}

This dataset contains $32$ classes with different numbers of samples per species.
We have selected $50$ images from each species (because that is the minimum number
of samples per species). $25$ of these are used as database images 
and $25$ are used as the testing images. We performed the same analysis as 
discussed above for the Swedish dataset. The results are shown in 
Figure \ref{flavia_pr}. Like the results for the Swedish dataset, we again 
significantly outperform state-of-the-art.

\subsection{Leafsnap dataset}
\begin{figure*}
\begin{center}
\begin{tabular}{c c}
\includegraphics[width=.48\textwidth]{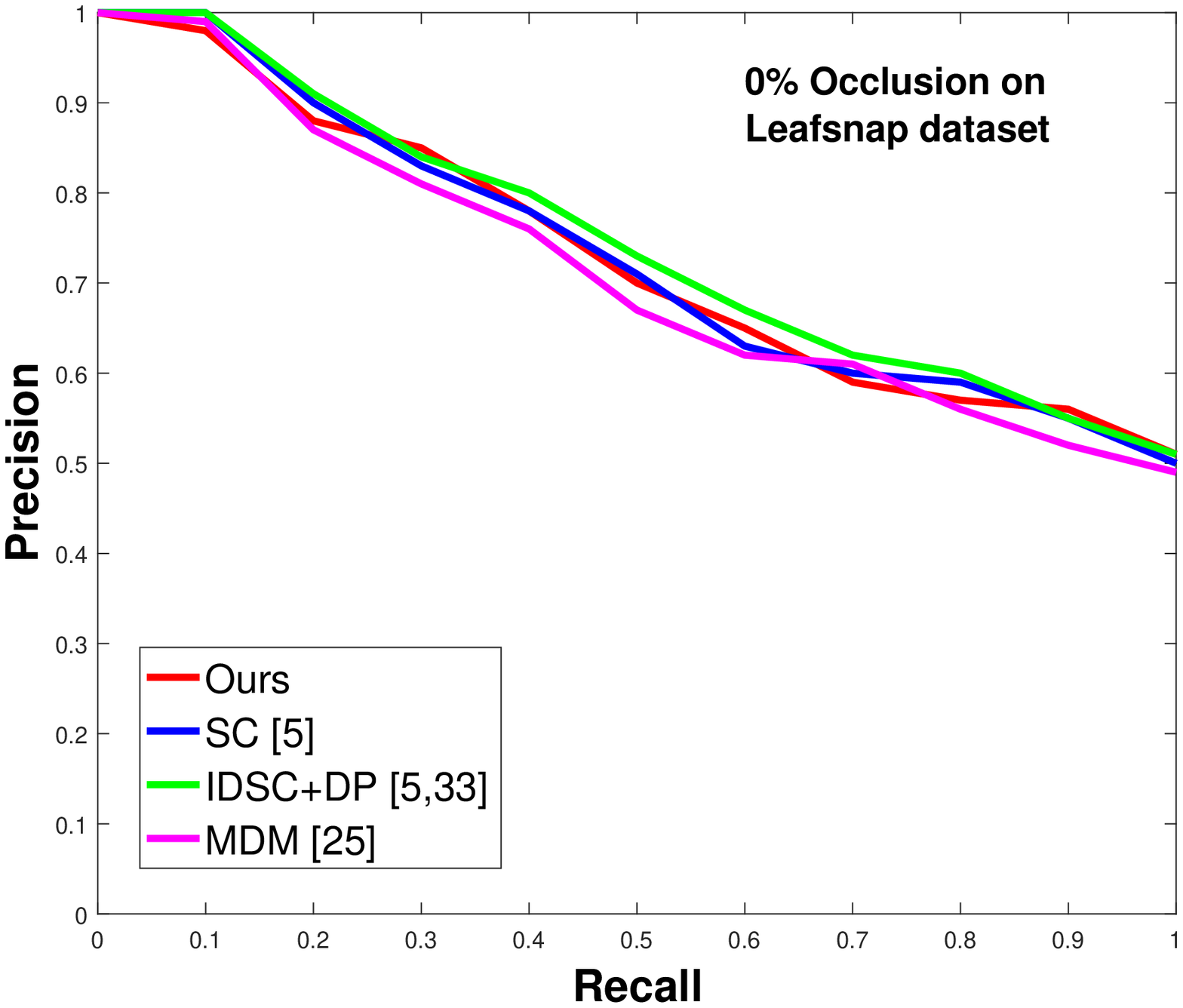} &
\includegraphics[width=.48\textwidth]{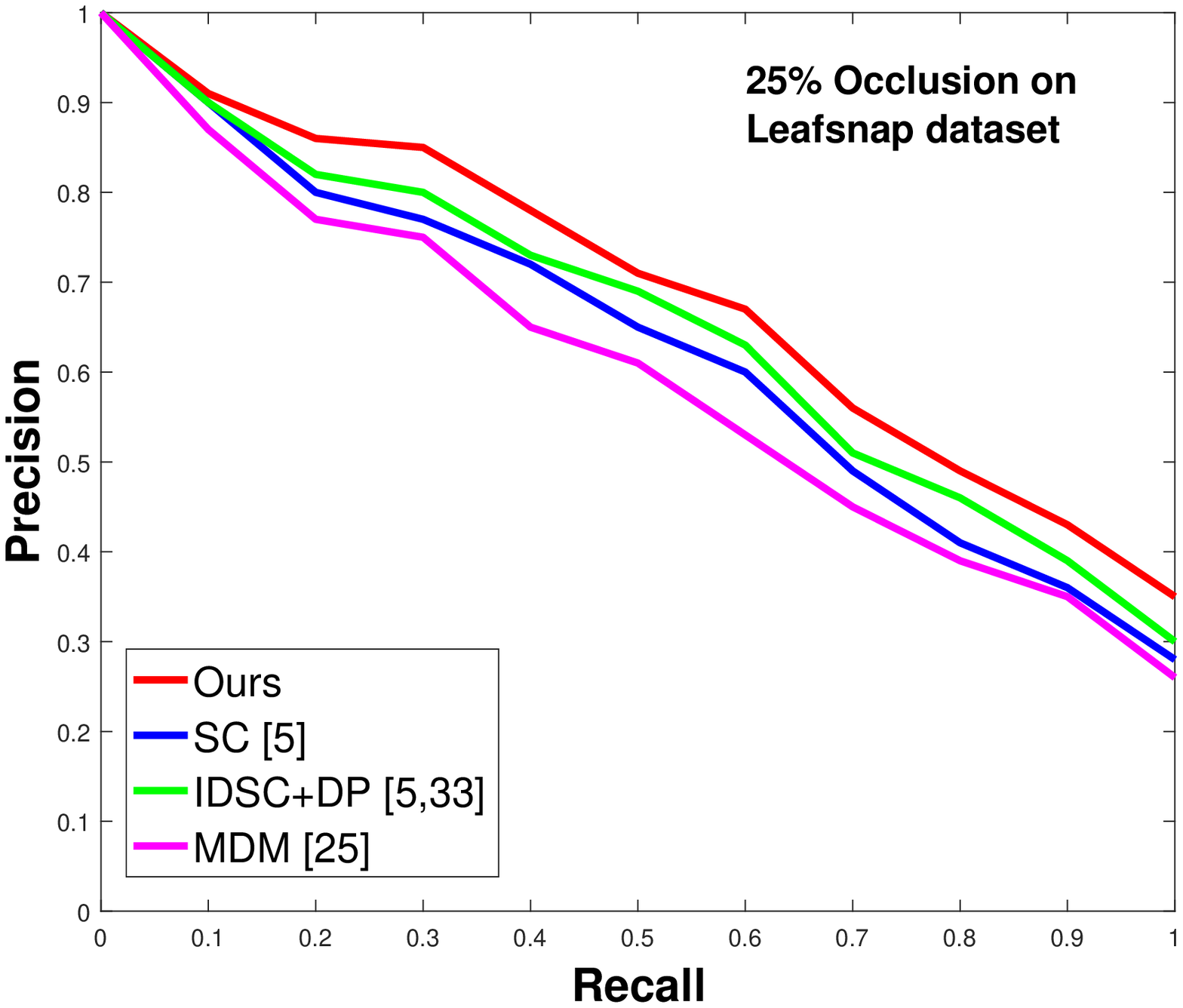} \\
(a) & (b) \\
\includegraphics[width=.48\textwidth]{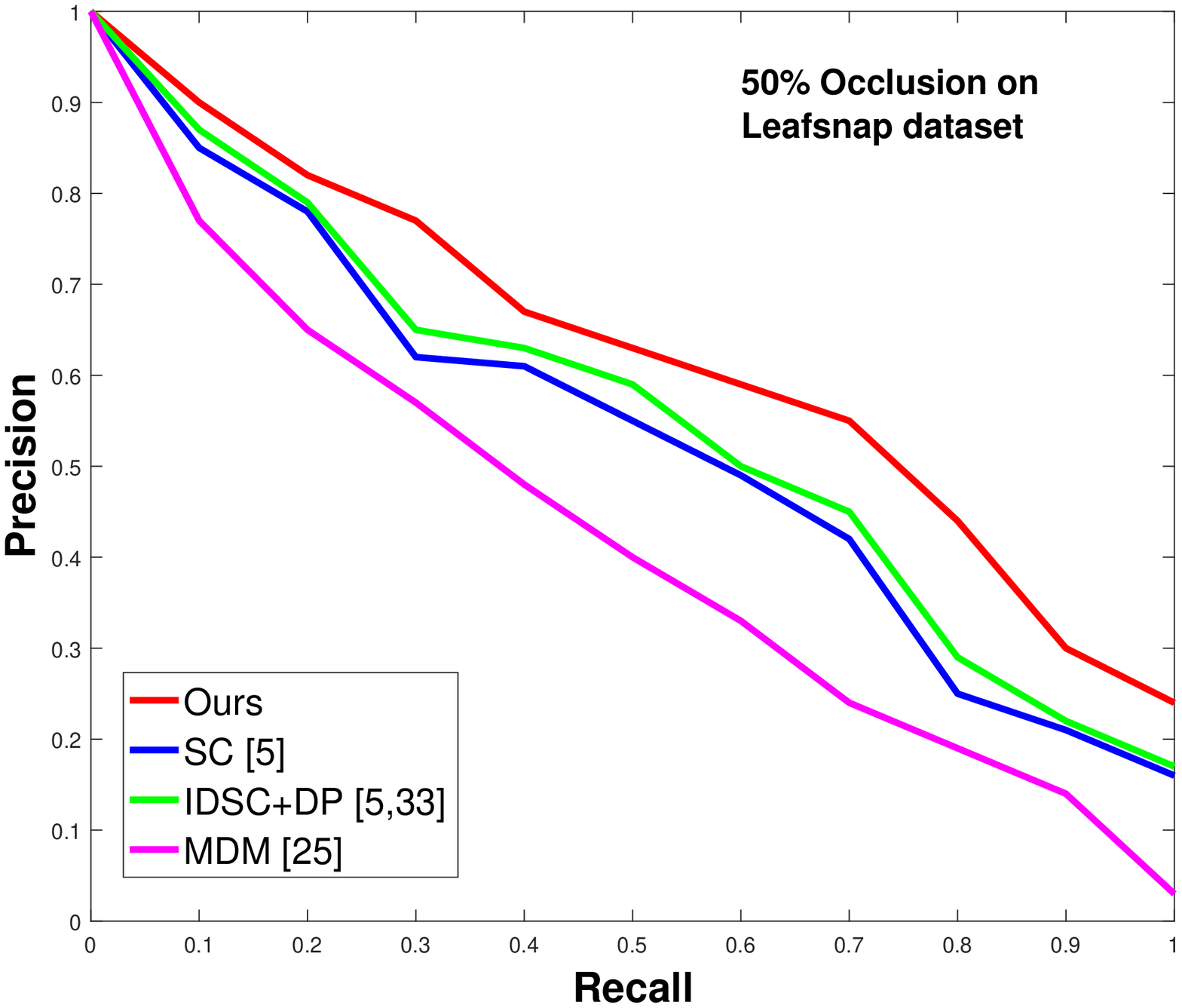} &
\includegraphics[width=.48\textwidth]{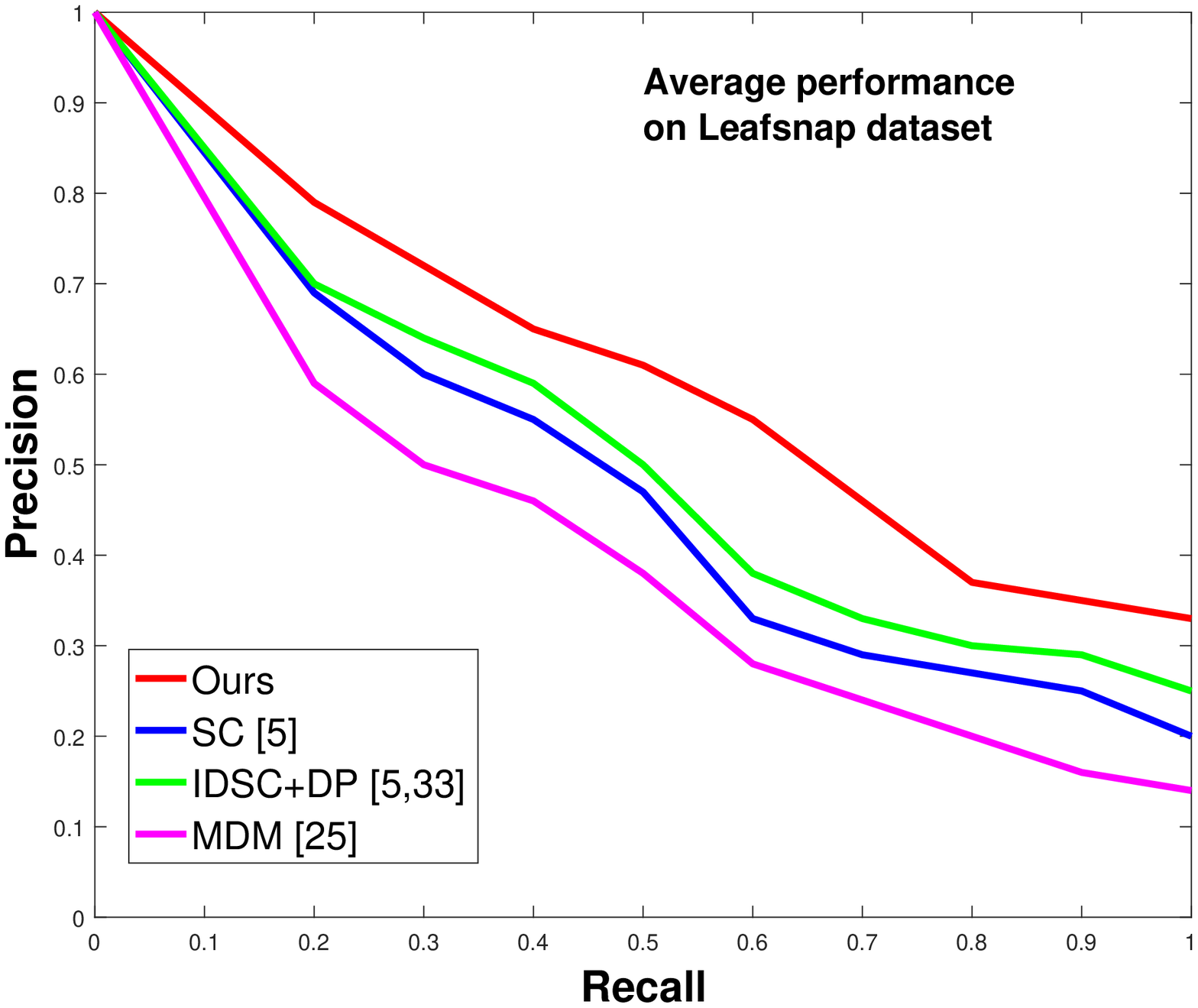}  \\
(c) & (d)
\end{tabular}
\end{center}
\caption{Precision-Recall curves for the Leafsnap dataset with (a) no occlusion, 
(b) $25\%$ occlusion, (c) $50\%$ occlusion and (d) average performance for random 
occlusion levels from $20\%$ to $50\%$ respectively.}
\label{leafsnap_pr}
\end{figure*}

This dataset contains $184$ species, having thousands of leaves in total. 
However, most of the species have a limited number of samples, which make them
unsuitable for testing. Moreover, due to segmentation errors, many samples are
not cleanly delimited. We also excluded compound leaf data from the testing.
Keeping all this in mind, we have selected 
a subset of $25$ species from the dataset. In each species, $20$ images are
used as database images and the rest
(depending on the number of available ``clean" images)
are used as the testing images. Results, shown in Figure \ref{leafsnap_pr}, 
are again significantly better
than state-of-the-art, as was the case for the Swedish and Flavia datasets.

\subsection{Discussion}
In all experiments, using all the different datasets, we outperform the state-of-the-art.
The average recognition rate for all the datasets is about $70.2\%$.
However, from the experimental results, it can be observed that 
our method performs the best for all occluded cases. When there is no occlusion,
IDSC \cite{IDSC-PAMI-2007} performs slightly better than our method
while our performance is quite close to SC \cite{SC-PAMI-2002}. 
In general, SC and IDSC perform similarly, while both outperform MDM \cite{MDM-TIP-2012}.
Although both SC and IDSC are widely used successfully on a large variety of shapes,
their performance drops for occluded leaves because both of the
methods perform point to point matching, and there is no way of
determining which part of a full curve best matches the occluded leaf.
A brute force approach would be to perform point-to-point
matching for all possible discrete combinations of the full curves in the 
database and find the best match, which is an NP-hard problem. 
One of the major contributions of our work is determining the section of the
full curve that closely resembles the occluded curve section. 

A drawback of the proposed method is that it is somewhat slow.
The optimization takes the majority of the time.
As stated in Zhao \emph{et al.} \cite{leafCounting-PR-2015}, the searching step is 
the main bottleneck in leaf recognition using
a large dataset. We use $n=1000$ points for the curves and the
GNCCP algorithm is slow because it is $O(n^3)$. Even using MatLab's \emph{parfor} to spawn 
separate computations for each of the 5 curves does not speed
things up enough.
Reducing the number of contour points is not a viable option as this
negatively affects the recognition accuracy. 

Lastly, note that our method is general
and can be used for other different applications of partial shape matching.

\section{Conclusion and Future Work}
We have presented an approach to recognize partially occluded leaves from
a database of different species.
Two immediate research directions would be to improve the recognition rate
and make the algorithm faster.
Also, we haven't investigated
compound leaves or the cases of classifying species
which are very close to each other. Using a leaf's texture as well as its contour
in the later case may improve classification results.

\nocite{*}
\bibliographystyle{IEEEannot}
\bibliography{ieee_tip}

\end{document}